\documentclass[10pt,twocolumn]{article}

\usepackage[letterpaper,top=0.72in,bottom=0.78in,left=0.72in,right=0.72in,columnsep=0.24in]{geometry}
\usepackage[T1]{fontenc}
\usepackage[utf8]{inputenc}
\usepackage{newtxtext,newtxmath}
\usepackage{courier}
\usepackage[hyphens]{url}
\usepackage[colorlinks=true,allcolors=blue!55!black]{hyperref}
\usepackage{xcolor}
\usepackage{graphicx}
\usepackage{booktabs}
\usepackage{multirow}
\usepackage{amsmath,amsfonts,mathtools}
\usepackage{algorithm}
\usepackage{algorithmic}
\usepackage{caption}
\usepackage{subcaption}
\usepackage{dblfloatfix}
\usepackage{newfloat}
\usepackage{listings}
\usepackage[numbers,sort&compress]{natbib}
\usepackage{microtype}

\DeclareCaptionStyle{ruled}{labelfont=normalfont,labelsep=colon,strut=off}
\floatstyle{ruled}
\newfloat{listing}{tb}{lst}{}
\floatname{listing}{Listing}
\lstdefinestyle{vlmprompt}{basicstyle=\ttfamily\footnotesize,columns=fullflexible,
  keepspaces=true,breaklines=true,breakatwhitespace=true,showstringspaces=false,
  numbers=none,frame=single,framerule=0.4pt,framesep=5pt,xleftmargin=0pt,
  xrightmargin=0pt,aboveskip=4pt,belowskip=4pt}

\title{\vspace{-0.45in}\textbf{FailPatch: Failure Residual Patching for\\Vision-Language-Action Models}}
\author{
Peng Yu\textsuperscript{1},
Jiacheng Wang\textsuperscript{1},
Ziheng Zhang\textsuperscript{2,\ensuremath{\dagger}},
Xuchong Zhang\textsuperscript{1,*},\\
Baoting Li\textsuperscript{1},
Zhuoyuan Yu\textsuperscript{2},
Yuxiang Chen\textsuperscript{3},
Tiancai Wang\textsuperscript{2},
Hongbin Sun\textsuperscript{1}\\[3pt]
\small \textsuperscript{1}Xi'an Jiaotong University \quad
\textsuperscript{2}Dexmal \quad
\textsuperscript{3}Nanjing University\\[-1pt]
\small \textsuperscript{*}Corresponding author \quad
\textsuperscript{\ensuremath{\dagger}}Project lead
}
\date{}

\begin{document}
\maketitle
\vspace{-0.10in}
\begin{abstract}
Vision-Language-Action (VLA) policies are typically adapted using successful demonstrations, which provide direct action supervision but rarely cover failure-prone states. Deployment failures expose these states, yet lack the corrective actions needed for conventional supervised learning. We propose \textbf{FailPatch}, a failure-driven residual patching framework that decouples action supervision from execution-reliability supervision. Successful demonstrations ground how the policy should act, while deployment trajectories indicate when its behavior becomes unreliable. We further observe that action hidden representations exhibit clear linear separability between reliable and failure-associated states while directly conditioning action generation. Building on these insights, FailPatch introduces a Null-gated Residual Expert Bank into the action hidden space of a frozen VLA policy. A unified \emph{Preserve--Redirect--Trust} objective retains the original policy in reliable states, selects residual experts in failure-associated states and redirects representations from failure regions toward success-associated regions under bounded intervention. With only 0.52\% trainable parameters, FailPatch improves success rates by 11.0 percentage points on four long-horizon RoboTwin tasks under clean evaluation, 9.5 percentage points under clean-to-random generalization, and 16.7 percentage points over the baseline across three real-world tasks. Project and code: \url{https://github.com/yupeng-2003/FailPatch}.
\end{abstract}


\section{Introduction}
\label{sec:introduction}

Vision-Language-Action (VLA) policies have emerged as a promising paradigm for robotic manipulation, mapping visual observations and language instructions directly to continuous actions~\cite{kim2024openvla,black2024pi0,li2024cogact, liu2024robomamba,liu2025rdt1b,wen2025dexvla, huang2025otter,li2025threedsvla}. These policies are commonly specialized through supervised fine-tuning on successful demonstrations. Although such demonstrations provide direct action supervision, they rarely cover the states reached after execution begins to deviate. In long-horizon tasks, small errors in grasping, placement, or physical interaction can propagate across action chunks and drive the policy into poorly represented states~\cite{fan2025longvla,zhao2025cotvla,zhong2026acotvla}. Improving deployment robustness therefore requires learning not only from successful behavior, but also from failures encountered during deployment.

\begin{figure}[!t]
    \centering
    \includegraphics[width=0.48\textwidth]{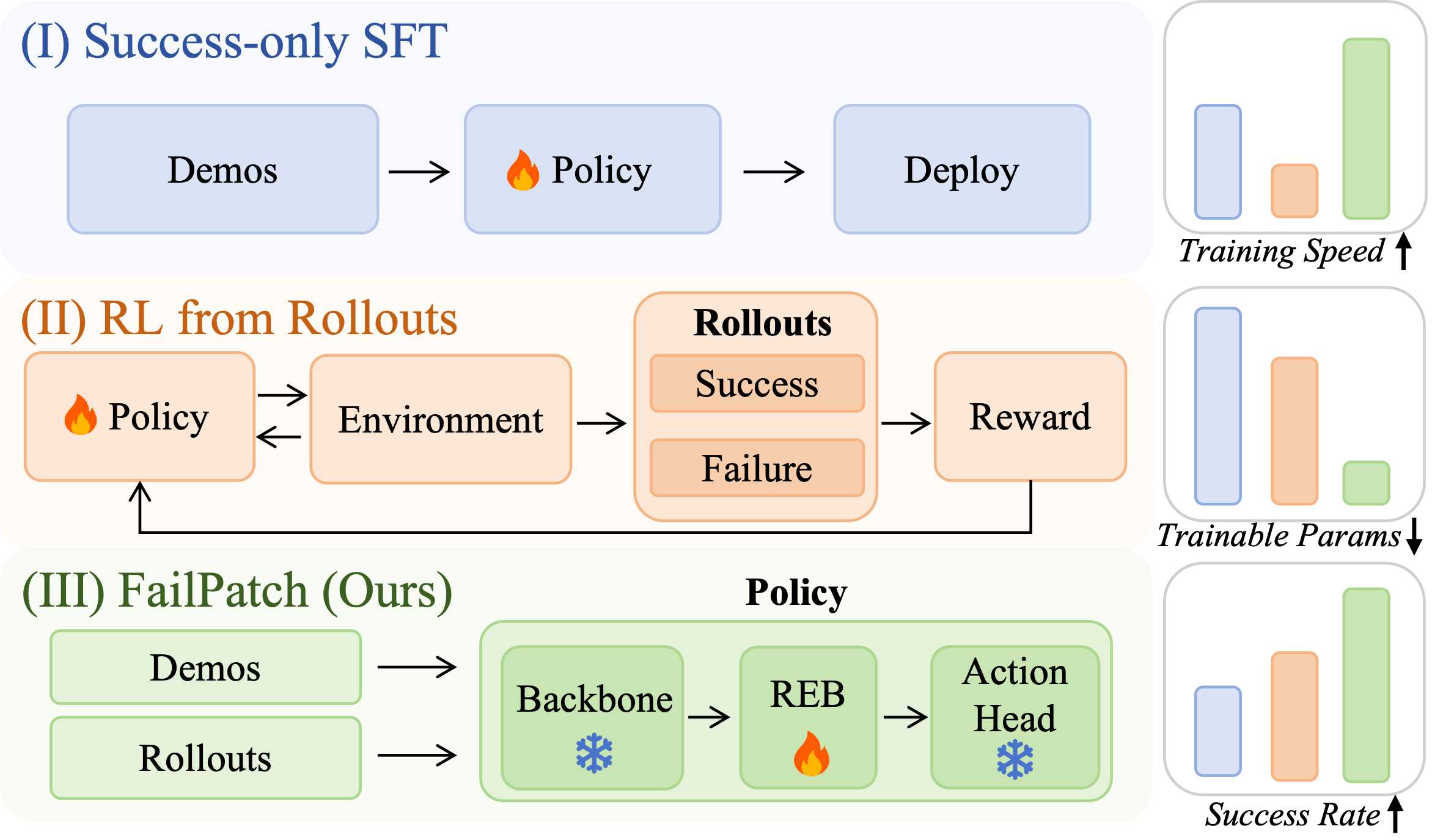}
    \caption{Comparison of VLA policy-learning paradigms. Success-only SFT learns from demonstrations, rollout-based RL optimizes from rewarded interaction, and FailPatch jointly exploits successful demonstrations and deployment rollouts through lightweight residual patching.}
    \label{fig1}
\vspace{-8pt}
\end{figure}

As illustrated in Fig.~\ref{fig1}, representative policy-improvement paradigms use deployment experience in fundamentally different ways. Success-only SFT learns action generation exclusively from successful demonstrations~\cite{kim2024openvla,black2024pi0}. Rollout-based reinforcement learning instead closes the deployment--learning loop by optimizing the policy from successful and failed interaction experience, but typically requires reward or value estimation, repeated environment interaction, and sometimes human corrections ~\cite{intelligence2025pi, pan2026sop,wang2026learning}. Beyond these policy-learning paradigms, a complementary line of work handles
failures through runtime detection or recovery~\cite{li2026tcot}. AHA~\cite{duan2025aha}, SAFE~\cite{gu2026safe}, and Hide-and-Seek~\cite{park2026hide} identify failure causes, likelihoods, or temporal signals, while FPC-VLA~\cite{yang2026fpc}, FailSafe~\cite{lin2025failsafe}, FLARE~\cite{zhao2026flare} and VLA-Corrector~\cite{pan2026vla} invoke external supervision, executable recovery actions, or online corrective replanning. However, an efficient supervised-learning framework that jointly exploits action-labeled demonstrations and outcome-only deployment failures remains underexplored, particularly when corrective-action annotations are unavailable and deployment should not rely on auxiliary failure detection or recovery modules.


We argue that successful demonstrations and deployment trajectories provide asymmetric but complementary information: demonstrations specify \emph{how} the policy should act, whereas deployment trajectories reveal \emph{when} its behavior becomes unreliable. Although failed trajectories cannot identify the corrective actions, they can supervise whether the original behavior should be preserved or redirected. We further observe that action hidden representations integrate visual and language context~\cite{liu2026ttfvla,li2026semanticvla,song2026reconvla} and exhibit clear linear separability between reliable and failure-associated states. As they directly participate in action generation, these representations are both outcome-sensitive and action-proximal, providing a natural interface for failure-driven residual patching.

Building on these insights, we propose \textbf{FailPatch}, an efficient failure-driven residual patching framework that decouples action supervision from execution-reliability supervision. Successful demonstrations train candidate residual corrections with executable action semantics, while deployment trajectories determine when the original policy should be retained or patched. An offline vision-language annotator localizes failure onset and filters reliable rollouts, enabling local success prototypes and failure-associated representations to provide geometric supervision. FailPatch instantiates this mechanism through a lightweight, Null-gated Residual Expert Bank (REB) inserted into the action hidden space of a frozen VLA policy. Through unified \emph{Preserve}, \emph{Redirect}, and \emph{Trust} constraints, the model preserves the original policy in reliable states, selects residual experts in failure-associated states, and redirects representations from failure regions toward success-associated representations with bounded intervention magnitude. The entire adaptation process requires no corrective-action annotations for failed trajectories, while online deployment requires neither an additional failure detector nor a replanning model.

We evaluate FailPatch on four long-horizon RoboTwin tasks and three real-world manipulation tasks. With only \(0.52\%\) trainable parameters (\(17.3\)M) and approximately 3 hours of training, FailPatch improves the average success rate over the baseline by \(11.0 \) percentage points under clean simulation evaluation and \(9.5 \) percentage points under clean-to-random generalization. On a physical dual-arm platform, it achieves an absolute improvement of 16.7 percentage points over the baseline across rigid-object, contact-rich, and deformable-object tasks. Ablations and mechanistic analyses further show that FailPatch preserves reliable execution, intervenes around failure onset, and redirects failure-associated representations toward success-associated representations.



	

\section{Related Work}
\label{sec:related_work}

\paragraph{Success-Supervised Robot Policy Learning.}
Generalist robot policies predominantly learn action generation from successful trajectories with explicit action targets. Early large-scale models, including RT-1~\cite{brohan2022rt}, RT-2~\cite{zitkovich2023rt}, and RT-X~\cite{o2024open}, demonstrated scalable robot-data training and cross-task or cross-embodiment transfer~\cite{zhao2023learning, chi2023diffusionpolicy}. More recent policies, such as Octo~\cite{team2024octo}, OpenVLA~\cite{kim2024openvla}, $\pi_0$~\cite{black2024pi0}, and $\pi_{0.5}$~\cite{black2025pi05}, further advance policy pretraining, multimodal co-training, and downstream fine-tuning~\cite{kim2025openvlaoft,pertsch2025fast,shukor2025smolvla, xu2025vlacache}. Although effective for acquiring demonstrated skills, success-supervised learning provides limited guidance for states reached after execution begins to deviate.

\paragraph{Learning from Deployment Experience and Failures.}
Recent work improves pretrained robot policies using experience collected during deployment~\cite{kelly2019hg, liu2025robot}. RECAP~\cite{intelligence2025pi} combines demonstrations, on-policy rollouts, and expert interventions through reinforcement learning, while SOP and Learning while Deploying scale online post-training across physical robot fleets~\cite{pan2026sop,wang2026learning}. AFIL more directly incorporates failure experience by using failed rollouts as negative guidance for action generation~\cite{zheng2026failing}. These approaches demonstrate the value of deployment data, but typically rely on reinforcement learning, reward or value estimation, expert interventions, or failure-conditioned action generation~\cite{zhang2025reinbot}. In contrast, FailPatch provides a lightweight supervised framework that jointly exploits successful demonstrations and deployment trajectories. It uses the former to learn action-grounded residual patches and the latter to supervise when the frozen policy should be preserved or patched, without corrective-action labels or auxiliary modules at deployment.

\section{Method}

\subsection{Problem Formulation and Overview}
\label{sec:problem_overview}

\paragraph{Problem formulation.}
Given a pretrained VLA policy \(\pi_{\phi}\), we consider offline residual patching using successful demonstrations \(\mathcal{D}_{\mathrm{demo}} =\{(\mathbf{o}_{i},\ell_{i},\mathbf{u}_{i})\}_{i=1}^{N_{\mathrm{demo}}}\), where \(\mathbf{o}_{i}\), \(\ell_i\), and \(\mathbf{u}_{i}\) denote the visual observation, language instruction, and action target, respectively. We also collect outcome-labeled deployment rollouts \(\mathcal{B}_{\mathrm{deploy}} =\{(\tau_i,y_i)\}_{i=1}^{N_{\mathrm{deploy}}}\), where \(\tau_i\) is generated by \(\pi_{\phi}\) and \(y_i\in\{0,1\}\) denotes its terminal outcome. Failed rollouts expose unreliable execution states but provide neither temporally localized failure labels nor corrective actions. Our goal is to learn lightweight residual parameters \(\theta\) while keeping \(\phi\) frozen, using successful demonstrations for action supervision and deployment rollouts for execution-reliability supervision.

\begin{figure*}[!t]
    \centering
    \includegraphics[width=0.95\textwidth]{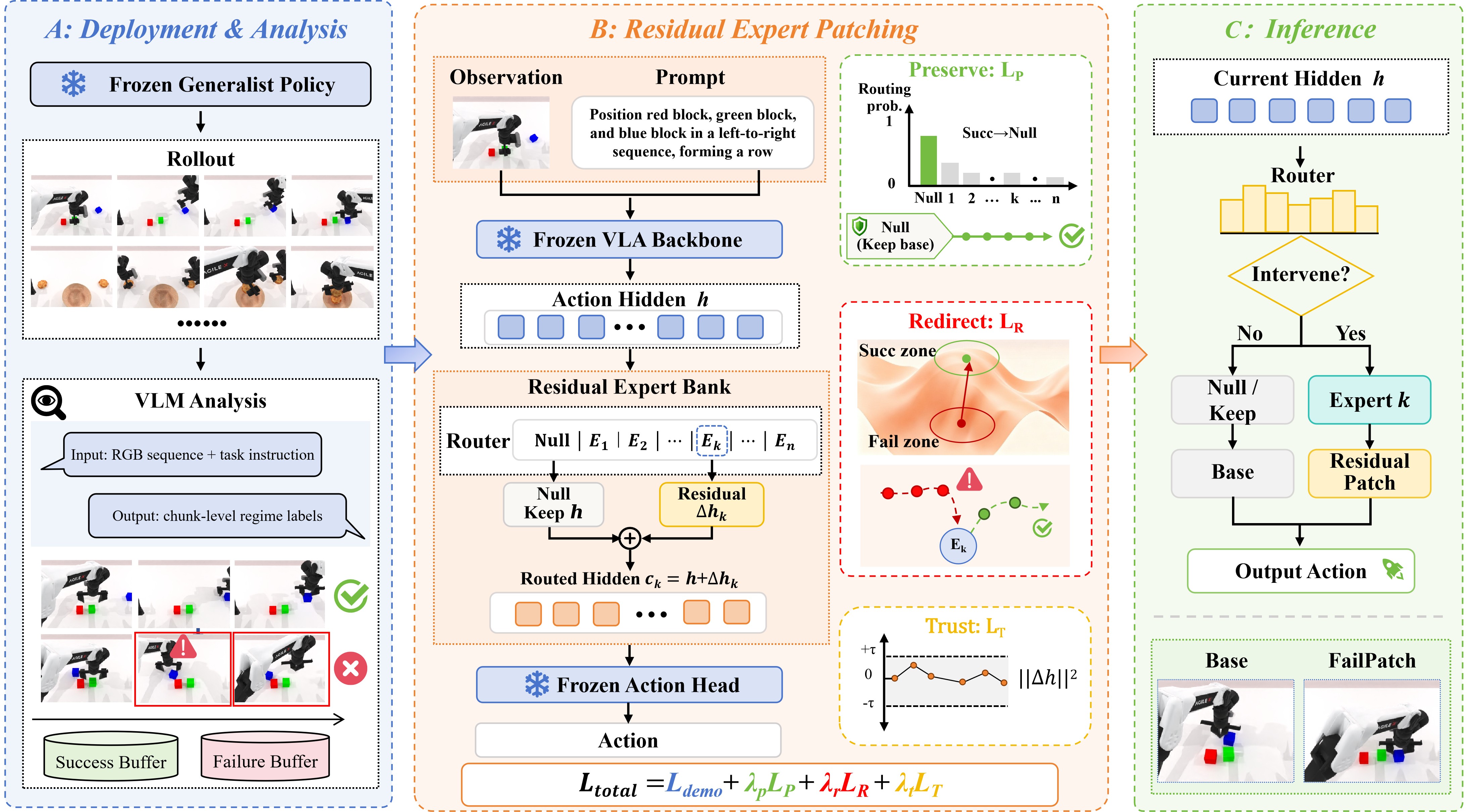}
    \caption{\textbf{Overview of FailPatch.}
    \textbf{(A)} A frozen VLA policy collects deployment rollouts, which an offline VLM converts into chunk-level reliability labels. \textbf{(B)} A Null-gated Residual Expert Bank is trained in the action hidden space using action supervision from successful demonstrations and \emph{Preserve--Redirect--Trust} constraints from deployment experience. \textbf{(C)} At inference, the router either retains the base representation through the Null expert or applies a selected residual patch.}
    \label{fig:main}
\vspace{-4pt}
\end{figure*}

\paragraph{Overview.}
As illustrated in Fig.~\ref{fig:main}, FailPatch is a failure-driven residual patching framework that decouples action supervision from execution-reliability supervision. First, the frozen policy is deployed to collect successful and failed rollouts, and an offline vision-language model produces chunk-level reliability labels and localizes failure onset. The resulting action hidden representations are organized into success- and failure-associated regions to provide temporally localized supervision. FailPatch then introduces a lightweight, Null-gated REB between the frozen VLA backbone and action head. The Null expert retains the original representation \(c_0=h\), whereas each residual expert produces a candidate patch \(\Delta h_k\), yielding \(c_k=h+\Delta h_k\). Successful demonstrations ground these patches in executable actions, while the \emph{Preserve}, \emph{Redirect}, and \emph{Trust} constraints supervise when to preserve the base policy, how to redirect failure-associated representations, and how strongly to intervene. The offline annotator and trajectory-derived supervision are discarded after training; inference retains only the frozen VLA and learned REB.

\subsection{Failure-Driven Residual Patching}
\label{sec:residual_routing}

\paragraph{Asymmetric supervision.}
FailPatch learns residual interventions from two asymmetric but complementary supervision sources. Successful demonstrations provide explicit action targets and therefore determine what constitutes an executable patch. Deployment rollouts provide no corrective actions, but reveal when the frozen policy becomes unreliable. FailPatch accordingly decouples patch learning from intervention supervision: demonstrations learn candidate action-grounded modifications, while deployment experience determines whether the original behavior should be preserved or redirected.

\vspace{-3pt}
\paragraph{Action-hidden intervention.}
Given a visual observation \(\mathbf{o}\) and language instruction \(\ell\), the frozen VLA produces an action hidden representation
\begin{equation}
\mathbf{h}
=
F_{\phi}(\mathbf{o},\ell),
\qquad
\mathbf{h}\in\mathbb{R}^{d}.
\label{eq:action_hidden}
\end{equation}
In our implementation, \(\mathbf{h}\) is obtained by mean-pooling the hidden states associated with the current action chunk. Using failure-onset-aligned rollouts, we find that a linear probe on the frozen action hidden representations can distinguish nominal pre-onset chunks from failure-associated onset and post-onset chunks. This observation indicates that execution reliability is already encoded in the action hidden space.

Meanwhile, \(\mathbf{h}\) lies immediately before the frozen action projection
$
\mathbf{v}_{0}
=
W_v\mathbf{h},
\label{eq:base_action}
$
and directly conditions the predicted action chunk. The action hidden space is therefore both \emph{outcome-sensitive} and \emph{action-proximal}: it exposes failure-related structure while providing a direct interface for modifying subsequent actions. FailPatch accordingly applies local residual patches to \(\mathbf{h}\), while keeping the pretrained VLA backbone and action projection frozen.

\paragraph{Sparse expert routing.}
A lightweight router predicts a categorical distribution over the Null and residual experts:
\begin{equation}
p_{\theta}(h)=\operatorname{softmax}(g_{\theta}(h)),
\qquad
p_{\theta}(h)\in\mathbb{R}^{K+1}.
\end{equation}
During training, all candidate expert routes are evaluated to construct separate action-grounding and router-supervision objectives, avoiding backpropagation through discrete routing decisions. At inference, we use sparse Top-1 routing:
\begin{equation}
r = \arg\max_{k\in\{0,\ldots,K\}} p_{\theta,k}(h),
\qquad
\tilde{h} = c_r .
\end{equation}

The selected representation is projected into the action space as \(v_{\theta}=W_v\widetilde{h}\). Selecting \(c_0\) preserves the base policy, whereas selecting \(c_k\) for \(k\geq 1\) applies the corresponding local residual patch. We use \(K=8\) residual experts and optimize only the router and experts, resulting in approximately \(17.3\)M trainable parameters.


\subsection{Preserve--Redirect--Trust Learning}
\label{sec:pet_objective}
FailPatch decouples the executable content of residual patches from their intervention supervision. Successful demonstrations provide direct action targets, while deployment trajectories determine which representations should be preserved or redirected. We instantiate this asymmetric supervision through the \emph{Preserve--Redirect--Trust} objective.
\paragraph{Localized trajectory supervision.}
Deployment rollouts provide only terminal outcomes, although failed trajectories may remain nominal before failure onset. We therefore use an offline vision-language annotator to localize \(\widehat{t}^{\,\mathrm{fail}}_i\) in each failed rollout and filter successful rollouts with a reliability indicator \(\gamma_i\). Specifically, we use GPT-5.5 as the offline annotator and provide it with the task instruction and temporally ordered RGB observations. For successful rollouts, \(\gamma_i=1\) only if no visually apparent intermediate deviation is identified. The implementation details are provided in the supplementary material. Across all 200 simulation failure rollouts, the VLM-localized onsets fall within $\pm1$ and $\pm2$ action chunks of manual annotations for 80\% and 95\% of the rollouts, respectively.

Let
\(\mathbf{h}_{i,t}=F_{\phi}(\mathbf{o}_{i,t},\ell_i)\) denote the action hidden representation at chunk \(t\). We partition deployment representations into reliable and failure-associated sets:
\vspace{0.5pt}
{\small
\begin{align}
\mathcal{H}^{+}
&=
\{\mathbf{h}_{i,t}\mid y_i=1,\gamma_i=1\}
\cup
\{\mathbf{h}_{i,t}\mid y_i=0,\,
t<\widehat{t}^{\,\mathrm{fail}}_i\},
\label{eq:reliable_set}\\
\mathcal{H}^{-}
&=
\{\mathbf{h}_{i,t}\mid y_i=0,\,
t\geq\widehat{t}^{\,\mathrm{fail}}_i\}.
\label{eq:failure_set}
\end{align}
}

Accordingly, reliable rollouts and nominal failure prefixes supervise preservation, whereas onset and post-onset states supervise redirection. Successful deployment rollouts supervise representations but are excluded from \(\mathcal{D}_{\mathrm{demo}}\).

For every failure-associated representation $h \in \mathcal{H}^{-}$, we retrieve its $K_{\mathrm{nn}}=20$ nearest neighbors from the task-specific pool of reliable representations extracted from successful rollouts using Euclidean distance $d(a,b)=\lVert a-b\rVert_2$, and define the corresponding local success prototype $s(h)$ as their coordinate-wise median.

\paragraph{Action-grounded patch learning.}
Successful demonstrations provide the only direct action supervision.
During training, we evaluate all residual-expert candidates rather than
backpropagating through discrete Top-1 routing. For each residual expert
\(k\in\{1,\ldots,K\}\), we define the candidate action loss as
\(\ell_k=\|W_v c_k-u\|_2^2\), and compute
\begin{equation}
\small
w_k=
\frac{e^{-\ell_k/T_{\mathrm d}}}
{\sum_{j=1}^{K}e^{-\ell_j/T_{\mathrm d}}},
\quad
\mathcal L_{\mathrm{expert}}
=
\mathbb E_{\mathcal D_{\mathrm{demo}}}\!
\left[\sum_{k=1}^{K}w_k\ell_k\right].
\end{equation}
Thus, every residual expert receives action supervision, with greater
weight assigned to candidates that better match the demonstrated action.


We additionally supervise the router with a demonstration-derived target.
The Null-candidate loss is
\(\ell_0=\|W_v c_0-u\|_2^2=\|W_v h-u\|_2^2\). We define
\begin{equation}
\small
q^{\mathrm{succ}}_0
=
\mathbb I\!\left[\min_{k\geq1}\ell_k\geq\ell_0\right],
\qquad
q^{\mathrm{succ}}_k
=
\left(1-q^{\mathrm{succ}}_0\right)w_k ,
\end{equation}such that the target selects Null when no residual candidate improves the base prediction and otherwise distributes responsibility according to the candidate action losses. The router is optimized by
\begin{equation}
\small
\mathcal L_{\mathrm{router\text{-}demo}}
=
\mathbb E_{\mathcal D_{\mathrm{demo}}}
\left[
\operatorname{CE}_{\mathrm{bal}}
\left(
\operatorname{sg}(q^{\mathrm{succ}}),
p_\theta(h)
\right)
\right].
\end{equation}
where $\operatorname{CE}_{\mathrm{bal}}$ denotes class-balanced cross-entropy and $sg(\cdot)$ denotes stop-gradient. 

The complete demonstration objective is
\begin{equation}
\small
\mathcal L_{\mathrm{demo}}
=
\mathcal L_{\mathrm{expert}}
+
\lambda_{\mathrm{rd}}
\mathcal L_{\mathrm{router\text{-}demo}} .
\end{equation}
This separates the optimization paths: the candidate action losses train
the residual experts, while the detached soft target trains the router.

\paragraph{Preserve reliable behavior.}
For \(\mathbf{h}\in\mathcal{H}^{+}\), intervention should be suppressed. We therefore favor the Null expert and penalize residual energy:
\begin{equation}
\mathcal{L}_P
=
E_{\mathcal{H}^+}
\big[
-\log p_{\theta,0}
+
\sum_k p_{\theta,k} \|\Delta \mathbf{h}_k\|^2
\big],
\label{eq:preserve_loss}
\end{equation}The first term learns abstention on reliable states, while the second limits unintended modifications by residual experts.

\begin{table*}[!t]
    \centering
    \small
    \vspace{-6pt}
    \begin{tabular}{lccccc}
        \toprule
        Method
        & Blocks Ranking
        & Stack Bowls
        & Put Object Cabinet
        & Place Bread Basket
        & Avg. \\
        \midrule
        \multicolumn{6}{l}{\textit{Without rollout trajectories}} \\
        50-Task Base
        & 38 & 68 & 40 & 52 & 49.50 \\
        LoRA SFT
        & 44 & 68 & 36 & 54 & 50.50 \\
        Full SFT
        & 44 & 70 & 42 & 55 & 52.75 \\
        Demo-Only REB
        & 42 & 64 & 36 & 55 & 49.25 \\
        \midrule
        \multicolumn{6}{l}{\textit{With rollout trajectories}} \\
        Outcome-Prompt SFT (Episode)
        & 39 & 51 & 44 & 29 & 40.75 \\
        Outcome-Prompt SFT (Chunk)
        & 42 & 70 & 35 & 37 & 46.00 \\
        PPO (RL)
        & 48 & 72 & 45 & 58 & 55.75 \\
        \textbf{FailPatch (Ours)}
        & \textbf{54}
        & \textbf{77}
        & \textbf{49}
        & \textbf{62}
        & \textbf{60.50} \\
        \bottomrule
    \end{tabular}
\caption{Success rates (\%) under clean evaluation.}
\label{tab:sim_clean}
\vspace{-3pt}
\end{table*}

\paragraph{Redirect failure-associated representations.}
For \(\mathbf{h}\in\mathcal{H}^{-}\), each residual expert produces a candidate \(\mathbf{c}_k\). We measure its utility by the progress toward the success prototype, regularized by patch magnitude:
\begin{equation}
a_k(\mathbf{h})
=
d(\mathbf{h},\mathbf{s(h)})
-
d(\mathbf{c}_k,\mathbf{s(h)})
-
\lambda_{\mathrm{u}}
\left\|
\Delta\mathbf{h}_k
\right\|_2^2.
\label{eq:routing_utility}
\end{equation}
The utilities define a soft routing target:
{\footnotesize
\begin{equation}
\begin{aligned}
q_0(\mathbf{h})
&=
\mathbb{I}
\left[
\max_{k\geq1}a_k(\mathbf{h})\leq m
\right],
\\
q_k(\mathbf{h})
&=
\left(1-q_0(\mathbf{h})\right)
\frac{
\exp\!\left(a_k(\mathbf{h})/T\right)
}{
\sum_{j=1}^{K}
\exp\!\left(a_j(\mathbf{h})/T\right)
}, k=1,\text{...},K.
\end{aligned}
\label{eq:routing_target}
\end{equation}
}The utility-derived distribution $\mathbf{q}(\mathbf{h})$ is treated as a stop-gradient target when optimizing the router. If no residual expert provides sufficient utility, the target retains the Null expert; otherwise, responsibility is distributed according to the relative candidate utilities.

Redirect aligns the learned router with this target and encourages selected patches to make margin-constrained progress toward the local success prototype:
{\small
\begin{equation}
\begin{split}
\mathcal{L}_R
=
E_{\mathcal{H}^-}
\Big[
& -\sum_j q_j \log p_{\theta,j} \\
& + \lambda_r \sum_k q_k \big[ m_r - (d(h,s) - d(c_k,s)) \big]_+
\Big],
\end{split}
\label{eq:redirect_loss}
\end{equation}
}The first term aligns the router with the utility-derived responsibilities, while the second enforces a minimum representational progress toward the local success prototype for selected residual experts. This objective uses failed trajectories to supervise intervention selection and representation redirection without constructing a failure prototype or inferring corrective actions.

\paragraph{Trust local intervention.}
Prototype-based gains could otherwise be achieved through excessively large hidden shifts. Trust therefore bounds the patches selected for failure-associated states:
\begin{equation}
\mathcal{L}_{\mathrm{T}}
=
\mathbb{E}_{\mathcal{H}^{-}}
\left[
\sum_{k}
q_k(\mathbf{h})
\left\|
\Delta\mathbf{h}_k
\right\|_2^2
\right].
\label{eq:trust_loss}
\end{equation}

The complete training objective is
\begin{equation}
\mathcal{L}_{\mathrm{total}}
=
\mathcal{L}_{\mathrm{demo}}
+
\lambda_{\mathrm{P}}\mathcal{L}_{\mathrm{P}}
+
\lambda_{\mathrm{R}}\mathcal{L}_{\mathrm{R}}
+
\lambda_{\mathrm{T}}\mathcal{L}_{\mathrm{T}}.
\label{eq:total_objective}
\end{equation}We use the first \(5\mathrm{k}\) of the \(30\mathrm{k}\)
training steps as a Null-disabled warm-up. During this period, the demonstration-derived routing target assigns probability only among the \(K\) residual experts, preventing the initially identical candidates
from collapsing to the Null route. Meanwhile, \(\mathcal L_{\mathrm{expert}}\) provides candidate-wise action supervision to all residual experts. After warm-up, the Null route is enabled when the best residual candidate does not improve over the base candidate according to the demonstration-routing criterion.

\section{Experiments}

\subsection{Simulation Experiments}
\label{sec:sim_experiments}

\paragraph{Setup.}
We evaluate FailPatch on four long-horizon RoboTwin~\cite{chen2025robotwin} tasks: Blocks Ranking, Stack Bowls, Put Object Cabinet, and Place Bread Basket, covering sequential multi-object manipulation, stacking, constrained placement, and object transport. We use \(\pi_{0.5}\)~\cite{black2025pi05} as the VLA policy. All methods start from a policy jointly pretrained on 50 manipulation tasks with 50 clean demonstrations per task, denoted the \emph{50-Task Base}. For each task, we reuse the same 50 demonstrations for adaptation and construct a fixed deployment buffer of 50 successful and 50 failed rollouts collected by the frozen base policy. All supervised adaptation methods are trained for \(30\mathrm{k}\) steps, while PPO is trained for \(100\) steps.


We compare against LoRA SFT~\cite{hu2022lora}, Full SFT, and \emph{Demo-Only REB}, a success-only variant with the same REB architecture as FailPatch. We further include two outcome-conditioned SFT baselines that append \emph{success: True/False} to the instruction. The episode-level variant assigns the terminal outcome to every action chunk, whereas the chunk-level variant uses temporally localized reliability labels. Both are evaluated under the desired condition \emph{success: True}. We additionally compare with PPO, which is optimized for 100 policy-update iterations using 128 on-policy rollouts per update. In contrast, FailPatch uses a fixed offline deployment buffer containing only 100 rollouts per task and requires no further environment interaction during adaptation. All adaptation runs are conducted on four NVIDIA H20 GPUs. Each reported result uses the same 100 environment seeds per task. Additional implementation details are provided in the supplementary material.


\paragraph{Main results.}
As shown in Table~\ref{tab:sim_clean}, FailPatch achieves the highest clean success rate on all four tasks, improving the average performance of the \emph{50-Task Base} from \(49.50\%\) to \(60.50\%\), an absolute gain of \(11.00\) percentage points. Its consistent advantage over success-only fine-tuning methods, including LoRA SFT, Full SFT, and Demo-Only REB, indicates that the improvement does not arise merely from continued training, increased adaptation capacity, or the REB architecture itself.

We further compare FailPatch with methods that use rollout trajectories. Episode-level Outcome-Prompt SFT reduces the average success rate to \(40.75\%\), below the \emph{50-Task Base}, showing that naively assigning terminal outcomes across an entire trajectory can degrade adaptation. Temporal localization recovers part of this loss, but the chunk-level variant still achieves only \(46.00\%\), indicating that directly conditioning action learning on reliability labels remains insufficient. With a comparable per-update rollout count, PPO achieves \(55.75\%\), while requiring 128 fresh on-policy rollouts per update and substantially longer training. These results demonstrate that FailPatch uses failure experience effectively by separating action supervision from execution-reliability supervision.

Under clean-to-random generalization (Table~\ref{tab:sim_random}), FailPatch achieves the highest average success rate of \(37.50\%\), improving the \emph{50-Task Base} by \(9.50\) percentage points. It performs best on three tasks and ties for the highest success rate on Stack Bowls, suggesting that residual patches learned from clean deployment failures transfer to unseen visual and physical variations.

\begin{table}[h]
    \centering
    \small
    \setlength{\tabcolsep}{1.2pt}
    \renewcommand{\arraystretch}{0.95}
    \begin{tabular}{@{}lccccc@{}}
        \toprule
        \multicolumn{1}{c}{Method}
        & \shortstack{Blocks\\Ranking}
        & \shortstack{Stack\\Bowls}
        & \shortstack{Object\\Cabinet}
        & \shortstack{Bread\\Basket}
        & Avg. \\
        \midrule
        50-Task Base
        & 12 & 39 & 25 & 36 & 28.00 \\

        LoRA SFT
        & 13 & 33 & 20 & 33 & 24.75 \\

        Full SFT
        & 23 & \textbf{45} & 29 & 34 & 32.75 \\

        Demo-Only REB
        & 20 & 39 & 31 & 32 & 30.50 \\

        Outcome SFT (Ep.)
        & 10 & 29 & 28 & 38 & 26.25 \\

        Outcome SFT (Ch.)
        & 16 & 35 & 32 & 32 & 28.75 \\
        \midrule

        \textbf{FailPatch}
        & \textbf{27}
        & \textbf{45}
        & \textbf{35}
        & \textbf{43}
        & \textbf{37.50} \\
        \bottomrule
    \end{tabular}
    \caption{Clean-to-random success rates (\%).}
    \label{tab:sim_random}
    \vspace{-6pt}
\end{table}


\begin{figure}[h]
    \centering
    \includegraphics[width=0.85\columnwidth]
    {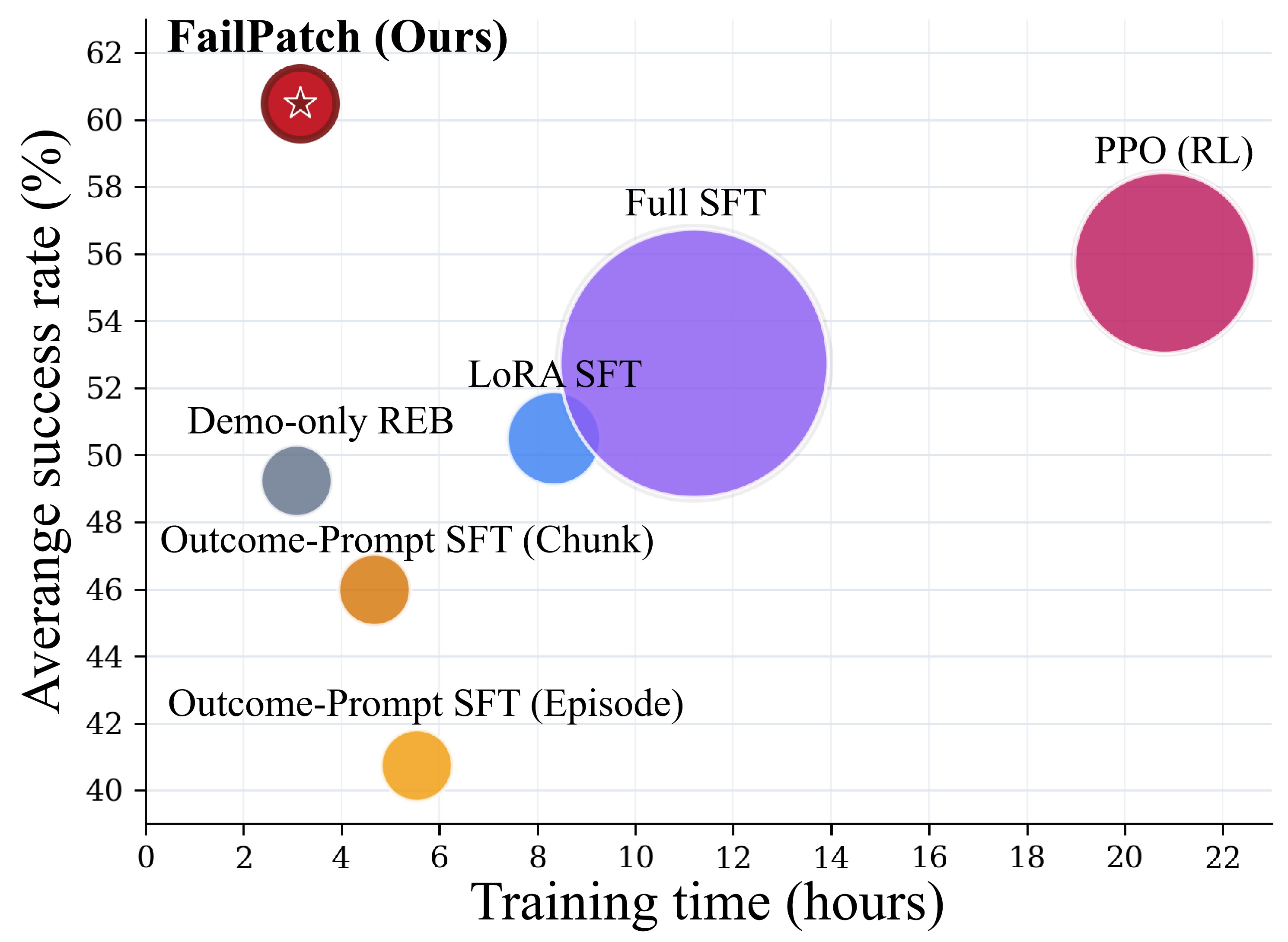}
    \vspace{-6pt}
    \caption{Adaptation performance and efficiency. Position denotes training time and average clean success rate, while \textbf{bubble size} represents the number of trainable parameters.}
    \label{fig:training_efficiency}
\end{figure}

\paragraph{Efficiency.}
Figure~\ref{fig:training_efficiency} compares adaptation performance,
trainable parameters, and training time. FailPatch trains only \(17.3\)M parameters (\(0.52\%\) of the VLA) in approximately \(3\) hours. Compared with LoRA SFT and Full SFT, it uses fewer trainable parameters and completes training \(2.6\times\) to \(3.6\times\) faster. PPO optimizes \(694\)M parameters and requires approximately \(20\) hours on the same hardware, together with repeated environment interaction. FailPatch nevertheless achieves a higher average success rate while using \(40.1\times\) fewer trainable parameters and completing training \(6 \times\) faster than PPO.

\subsection{Real-World Experiments}
\label{sec:real_experiments}

\begin{figure}[h]
    \centering
    \includegraphics[width=0.88\columnwidth]{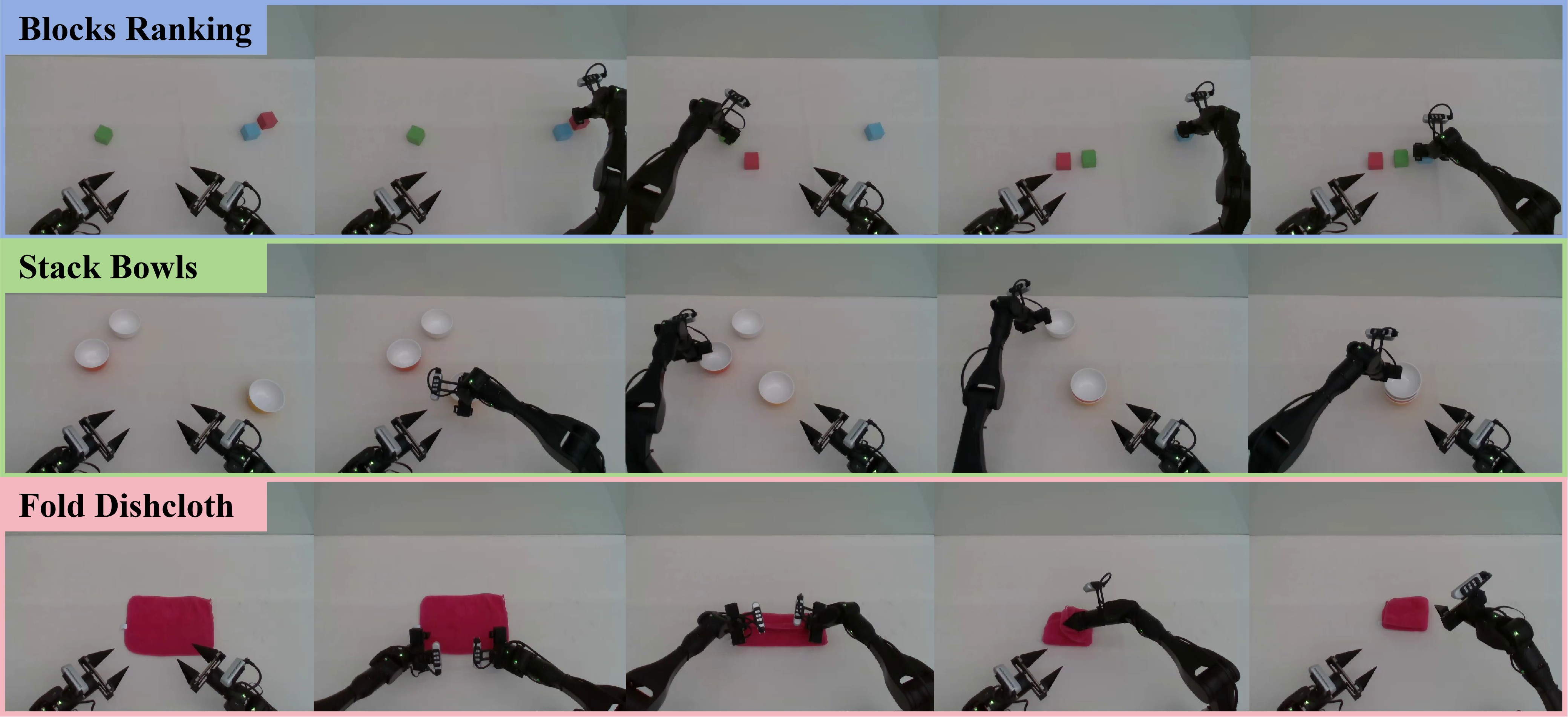}
    \caption{Illustrations of the three real-world manipulation tasks. From top to bottom: Blocks Ranking, Stack Bowls, and Fold Dishcloth.}
    \label{fig:real_world_setup}
\vspace{-8pt}
\end{figure}

\paragraph{Setup.}
We evaluate FailPatch on three real-world tasks: Blocks Ranking, Stack Bowls, and Fold Dishcloth, which respectively involve sequential multi-object manipulation, contact-rich stacking, and deformable-object control. Experiments are conducted on a dual-arm platform equipped with an Intel RealSense D435i camera. Figure~\ref{fig:real_world_setup} shows the three tasks from top to bottom. For each task, we collect 100 successful demonstrations and a deployment buffer containing 50 successful and 50 failed rollouts from the base policy. All methods start from the same \(\pi_{0.5}\) VLA policy ~\cite{black2025pi05}, jointly trained on the three tasks and referred to as the \emph{Three-Task Base}. All supervised adaptation methods are trained for \(30\mathrm{k}\) optimization steps.



\paragraph{Results.}
As shown in Table~\ref{tab:real_results}, FailPatch achieves the highest success rate on all three tasks, improving the average performance of the \emph{Three-Task Base} from \(41.1\%\) to \(57.8\%\), an absolute gain of \(16.7\) percentage points. Improvements are observed across all three manipulation settings rather than being concentrated on a single task, suggesting that rollout-derived reliability supervision can support intervention under distinct real-world failure modes, including accumulated ordering errors, unstable object interactions, and deformation-dependent execution deviations.

\begin{table}[t]
    \centering
    \normalsize
    \setlength{\tabcolsep}{2.5pt}
    \begin{tabular}{@{}lcccc@{}}
        \toprule
        \multicolumn{1}{c}{\multirow[c]{2}{*}{Method}}
        & Blocks
        & Stack
        & Fold
        & \multirow[c]{2}{*}{Avg.} \\
        & Ranking
        & Bowls
        & Dishcloth
        & \\
        \midrule
        Three-Task Base
        & 36.7
        & 53.3
        & 33.3
        & 41.1 \\
        LoRA SFT
        & 46.7
        & 50.0
        & 40.0
        & 45.6 \\
        Demo-Only REB
        & 43.3
        & 50.0
        & 36.7
        & 43.3 \\

        \textbf{FailPatch}
        & \textbf{56.7}
        & \textbf{66.7}
        & \textbf{50.0}
        & \textbf{57.8} \\
        \bottomrule
    \end{tabular}
    \caption{Real-world success rates (\%) over 30 trials per task.}
    \label{tab:real_results}
\vspace{-10pt}
\end{table}

\subsection{Ablation Studies}
\label{sec:ablation}

We report the objective and architecture ablations in Table~\ref{tab:ablation}. Additional ablations on prototype construction and hyperparameter sensitivity are provided in the supplementary material. Objective variants add the indicated constraints to Demo-Only REB, whereas architecture variants retain the full objective. Redirect provides the largest standalone gain, confirming that failure-associated representations offer useful supervision for residual intervention. Preserve complements Redirect by retaining reliable base behavior, while adding Trust to both constraints yields the full performance of 60.50\%. Replacing the expert bank with a parameter-matched single expert or removing the Null expert reduces success, supporting the need for diverse patch directions and learned abstention. Using manually annotated failure onsets yields comparable performance to the VLM-annotated variant, suggesting that the automatic annotations provide near-oracle supervision despite modest localization errors.


\begin{table}[h]
    \centering
    \small
    \vspace{-6pt}
    \begin{tabular}{lc}
        \toprule
        Variant & Avg. Success (\%) \\
        \midrule
        \multicolumn{2}{l}{\textit{Objective components}} \\
        Demo-Only REB
        & 49.25 \\
        \quad + Preserve
        & 52.50 \\
        \quad + Redirect
        & 56.50 \\
        \quad + Trust
        & 50.25 \\
        \quad + Preserve + Redirect
        & 59.00 \\
        \midrule
        \multicolumn{2}{l}{\textit{Architecture}} \\
        Parameter-Matched Single Expert
        & 55.75 \\
        REB without Null Expert
        & 56.50 \\
        \midrule
        \emph{Annotation} \\
        Human-Annotated Onset & 61.00 \\
        \midrule
        \textbf{Full FailPatch}
        & \textbf{60.50} \\
        \bottomrule
    \end{tabular}
\caption{Ablation results averaged across simulation tasks.}
\label{tab:ablation}
\vspace{-8pt}
\end{table}

\subsection{Mechanistic Analysis}
\label{sec:mechanistic_analysis}

\begin{figure}[t]
    \centering
    \includegraphics[width=0.68\columnwidth]{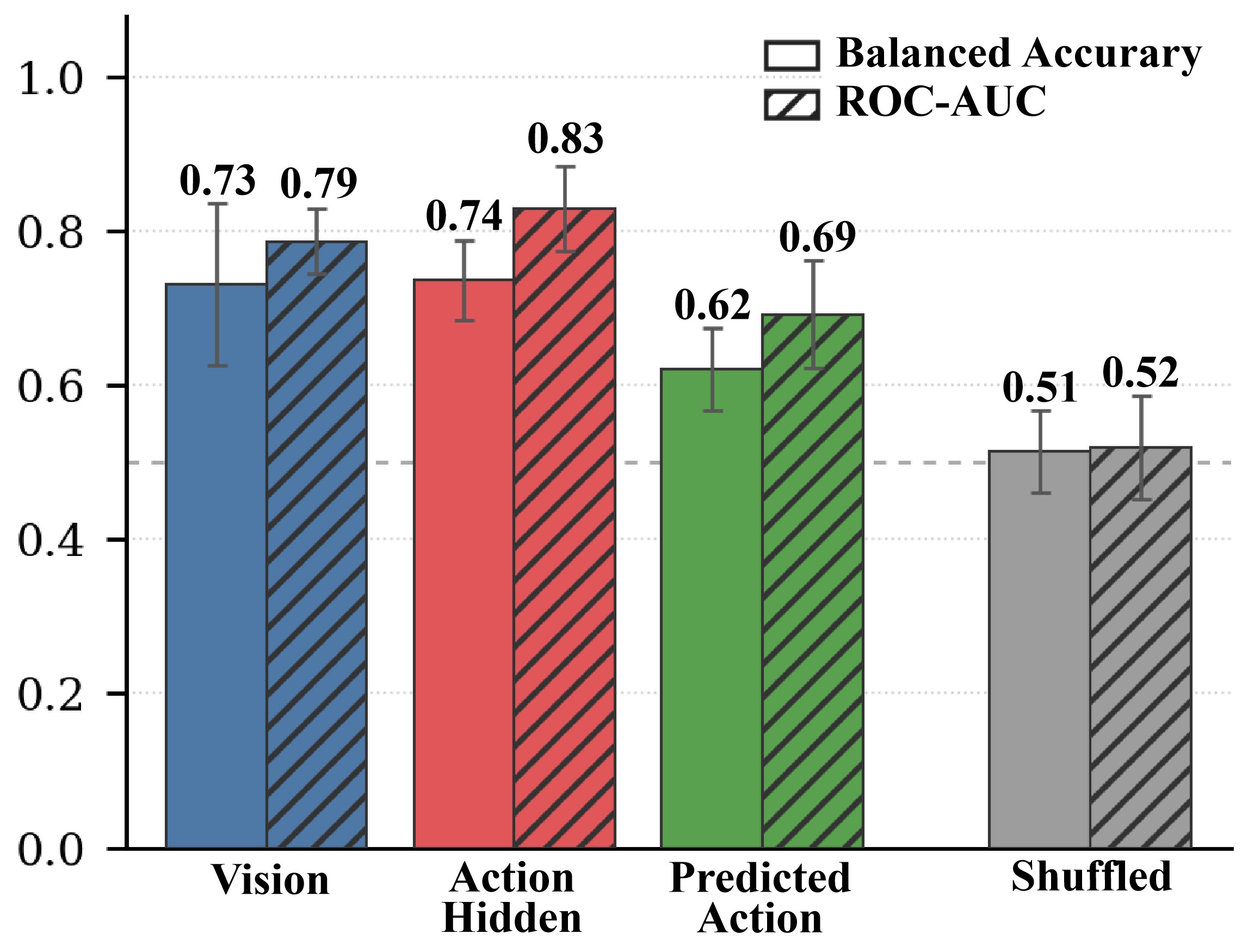}
    \vspace{-7pt}
    \caption{Linear separability of nominal and failure-associated chunks across representation spaces. Shuffled labels serve as a chance-level control.}
    \label{fig:linear_probe}
    \vspace{-10pt}
\end{figure}

\paragraph{Why intervene in the action hidden space?}
We compare frozen visual features, action hidden representations, and predicted actions using linear probes that distinguish nominal pre-onset chunks from failure-associated chunks, with rollout-level splits to avoid trajectory leakage. As shown in Fig.~\ref{fig:linear_probe}, action hidden representations achieve the strongest separability, with a balanced accuracy of \(0.74\) and a ROC-AUC of \(0.83\), while shuffled-label controls remain near chance. Since this space both encodes execution reliability and directly conditions action generation, it provides a suitable interface for residual intervention.

\begin{figure}[h]
    \centering
    \vspace{-4pt}
    \begin{minipage}[t]{0.49\linewidth}
        \centering
        \includegraphics[width=\linewidth]{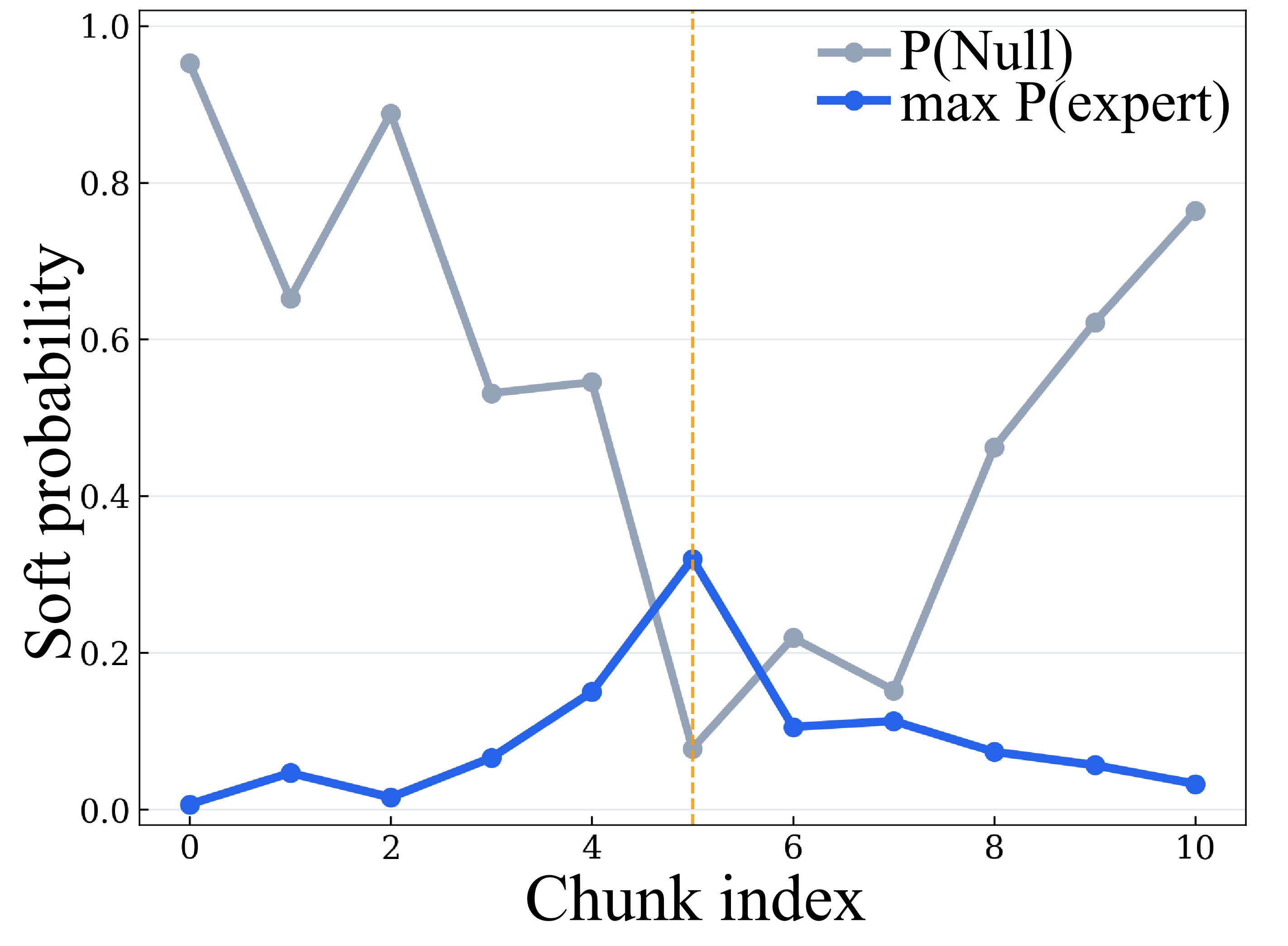}\\[-2pt]
    \end{minipage}
    \hfill
    \begin{minipage}[t]{0.49\linewidth}
        \centering
        \includegraphics[width=\linewidth]{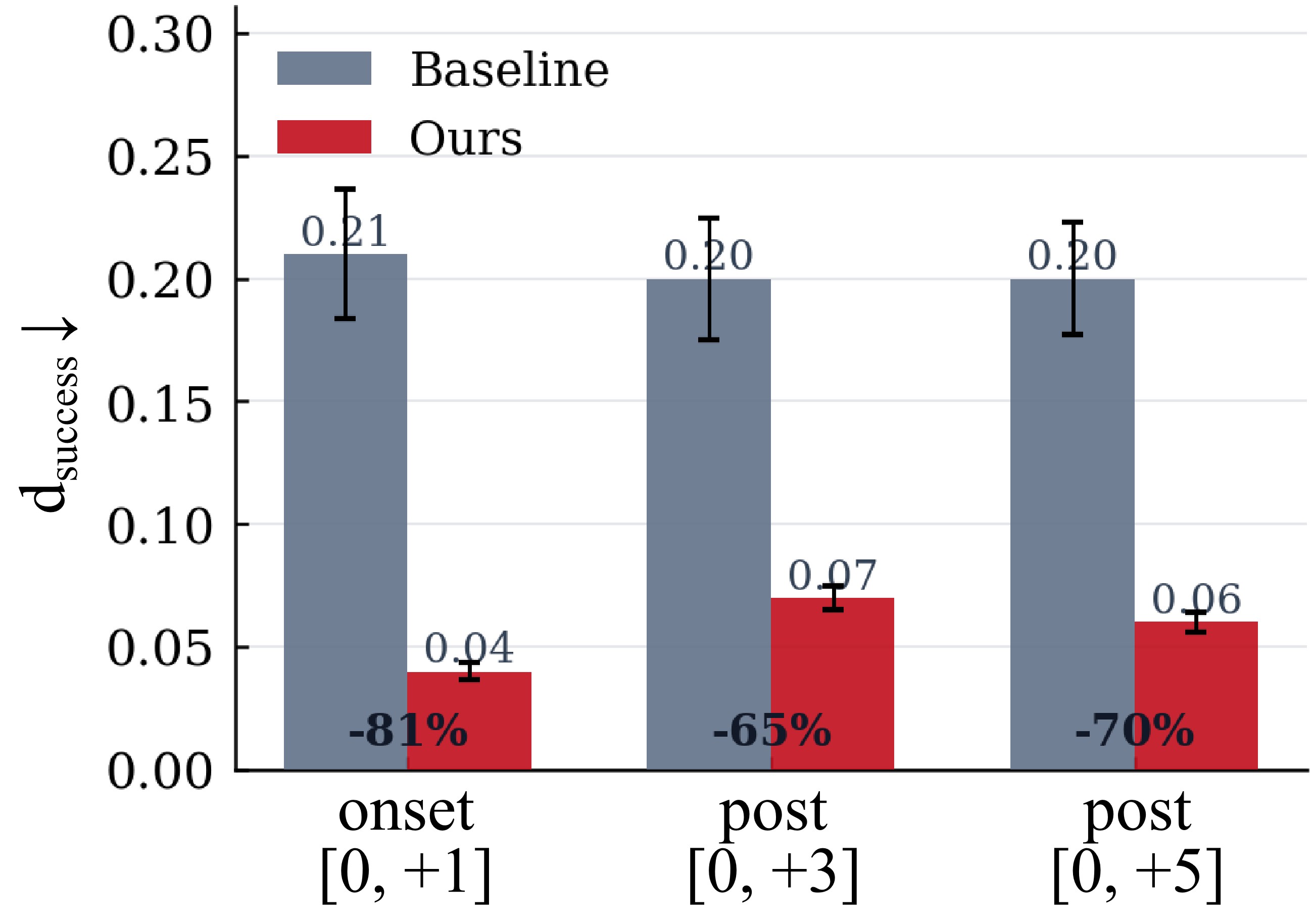}\\[-2pt]
    \end{minipage}
    \vspace{-4pt}
    \caption{(a) Soft routing probabilities along a representative episode. (b) Distance to the local success prototype before and after patching.}
    \label{fig:intervention_geometry}
    \vspace{-8pt}
\end{figure}

\paragraph{When does FailPatch intervene?}
Figure~\ref{fig:intervention_geometry}(a) shows the soft routing probabilities along a representative episode. The Null route dominates most action chunks, but its probability drops sharply near the localized failure onset, where the highest residual-expert probability peaks and triggers intervention. The router subsequently returns toward Null, indicating a localized rather than persistent correction.

\paragraph{How does patching reshape failure-associated representations?}
Figure~\ref{fig:intervention_geometry}(b) provides a geometric diagnostic of the selected patches. Around failure onset, patching substantially reduces the distance to the local success prototype, and this effect persists over subsequent action chunks. This confirms that FailPatch redirects failure-associated representations toward reliable regions.


\begin{figure}[h]
    \centering
    \begin{minipage}[t]{0.45\linewidth}
        \centering
        \includegraphics[width=\linewidth]{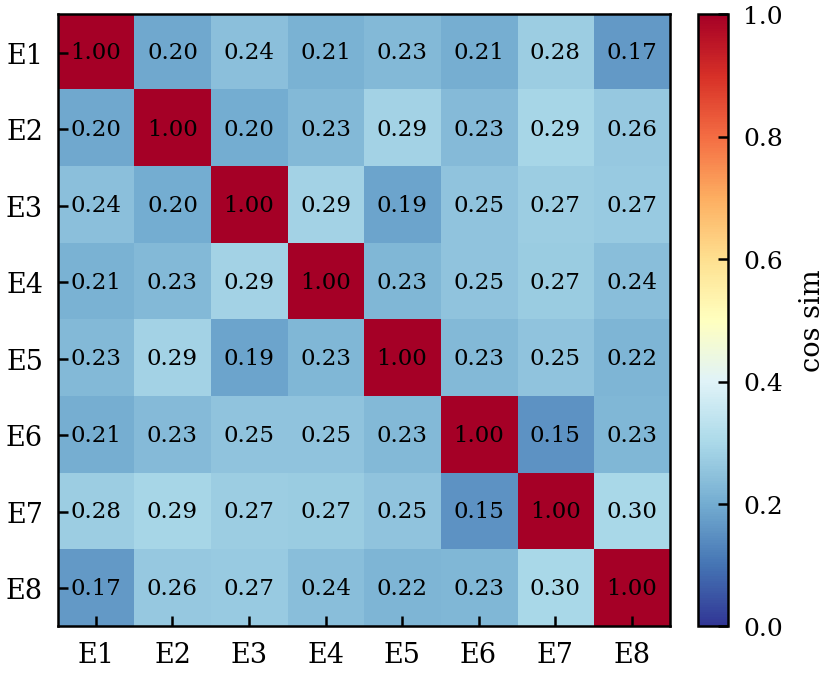}\\[-2pt]
        {\small (a) Hidden residuals \(\Delta h_k\)}
    \end{minipage}
    \hfill
    \begin{minipage}[t]{0.45\linewidth}
        \centering
        \includegraphics[width=\linewidth]{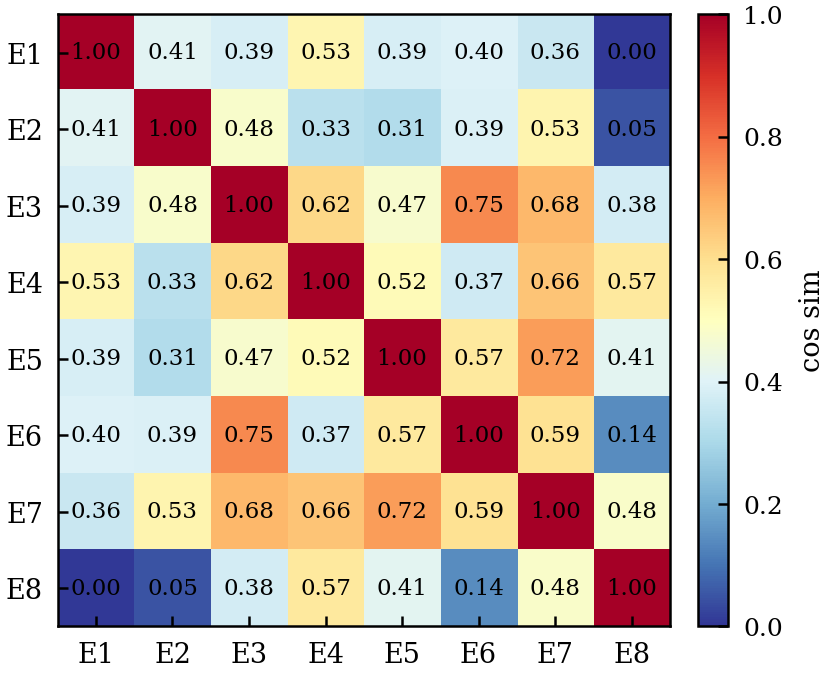}\\[-2pt]
        {\small (b) Action residuals \(W_v\Delta h_k\)}
    \end{minipage}
    \vspace{-4pt}
    \caption{Residual-expert diversity at localized failure onsets.}
    \label{fig:expert_diversity}
    \vspace{-8pt}
\end{figure}

\paragraph{Do residual experts learn distinct intervention directions?}
At each localized failure onset, we pass the same hidden representation through all residual experts and compute pairwise cosine similarities between their hidden residuals \(\Delta h_k\) and projected action residuals \(W_v\Delta h_k\). Figure~\ref{fig:expert_diversity} shows low off-diagonal similarities in both spaces, indicating that the experts learn distinct intervention directions rather than collapsing to similar behaviors.


\begin{figure}[h]
    \centering
    \includegraphics[width=\columnwidth]{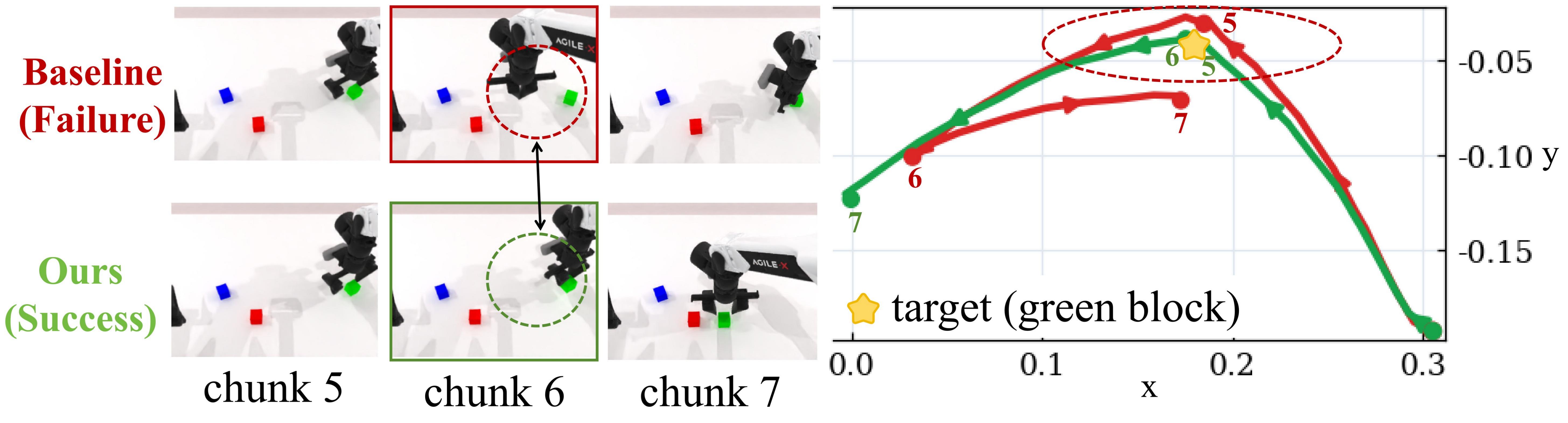}
    \vspace{-14pt}
    \caption{Representative Blocks Ranking rescue episode. Left: chunks 5--7 from the same initial configuration. Right: 2D TCP trajectories, where FailPatch makes a small local correction near the green block.}
    \label{fig:vis}
    \vspace{-12pt}
\end{figure}

\paragraph{What does FailPatch correct in practice?}
Figure~\ref{fig:vis} shows a representative rescue episode from the same initial configuration. The left panel visualizes chunks 5--7 for the baseline and FailPatch. At chunk 6, the frozen base policy makes a slightly misaligned approach to the green block and subsequently fails to grasp it, while FailPatch applies a local correction at the same stage and successfully reaches the target, leading to task completion. The 2D TCP projection on the right further shows that the successful rescue requires only a small trajectory adjustment in the highlighted region near the target block, rather than a large global deviation. This demonstrates that a localized hidden-space intervention can translate into a decisive corrective action during execution.
\vspace{-4pt}

\section{Conclusion}
We propose FailPatch, an efficient failure-driven residual patching framework for adapting pretrained VLA policies from deployment experience. It decouples action supervision from execution-reliability supervision: successful demonstrations ground executable residual patches, while deployment trajectories determine when the frozen policy should be preserved or corrected. A Null-gated Residual Expert Bank applies these patches in the action hidden space through the Preserve--Redirect--Trust objective, without corrective-action labels or auxiliary modules at deployment. Experiments in simulation and on real robots demonstrate performance gains with high parameter efficiency and substantially reduced training time. FailPatch relies on offline VLM-based failure-onset annotations; reducing this dependence through self-supervised or uncertainty-aware localization remains future work.

\clearpage
\bibliographystyle{unsrtnat}
\bibliography{references}

\clearpage
\appendix
\setcounter{section}{0}
\renewcommand{\thesection}{S\arabic{section}}
\renewcommand{\thesubsection}{S\arabic{section}.\arabic{subsection}}
\section*{Supplementary Material}
\addcontentsline{toc}{section}{Supplementary Material}

\section*{Overview}

This supplement provides implementation and evaluation details omitted from the main paper. Additional Method Details describes the frozen $\pi_{0.5}$ policy, Residual Expert Bank, local success prototypes, optimization settings, and the complete training and inference procedure. Offline VLM Annotation presents the annotation inputs, verbatim prompts, reliability filtering, and agreement with human annotations. Experimental Details reports the simulation horizons, success predicates, randomization settings, domain shifts, baseline configurations, and compute cost. Additional Ablation Studies provides further analyses of prototype construction, failure-data efficiency, hyperparameters, objective components, and model architecture, while Additional Mechanistic Analysis details the linear-probe protocol. Real-World Experimental Details describes the physical platform, tasks, and evaluation protocol. Finally, the Limitations section discusses the primary limitation of FailPatch.

\section{Additional Method Details}
\label{supp:sec:method}

\subsection{Backbone and Action-Hidden Interface}

We implement FailPatch on top of the OpenPI implementation of
$\pi_{0.5}$~\citep{black2025pi05}.  The simulation policy receives three RGB observations,
comprising one scene view and two wrist views, together with the original
task instruction.  The images are resized to $224\times224$ and processed
using the same visual preprocessing as the pretrained policy.  We reuse the
normalization statistics of the 50-task base policy for every adaptation
method and do not recompute task-specific action statistics.

Let the action horizon be $H=50$.  Immediately before the frozen action
projection, the policy produces action-token hidden states
$Z\in\mathbb{R}^{H\times d}$, where $d=1024$.  We form the chunk-level
representation used by the router as
\begin{equation}
    h=\frac{1}{H}\sum_{t=1}^{H}Z_t.
\end{equation}
The residual expert selected for the chunk is applied to all action-token
hidden states through the same action-hidden interface.  The resulting hidden
states are passed to the frozen action projection, whose model output width is
32.  The backbone, visual encoder, language model, action expert, and action
projection remain frozen throughout FailPatch adaptation.

\subsection{Residual Expert Bank}

The router first applies LayerNorm to the pooled representation and then uses
a single linear layer to predict $K+1$ logits:
\begin{equation}
    p_{\theta}(h)
    =
    \operatorname{softmax}\!\left(g_{\theta}(
    \operatorname{LN}(h))\right),
    \qquad
    g_{\theta}:\mathbb{R}^{1024}\rightarrow\mathbb{R}^{9}.
\end{equation}
The nine routes comprise one Null route and $K=8$ residual experts. Its bias is initialized to zero.
During training, the complete routing distribution supervises expert
responsibilities.  During execution, FailPatch uses sparse Top-1 routing.

Each residual expert is a SwiGLU bottleneck MLP.  With bottleneck width
$d_e=704$, expert $k$ computes
\begin{equation}
    \Delta h_k
    =
    W_{k,\mathrm{down}}
    \left[
    \operatorname{SiLU}(W_{k,\mathrm{gate}}h)
    \odot
    (W_{k,\mathrm{up}}h)
    \right],
\end{equation}
where the gate and up projections map from 1024 to 704 dimensions and the
down projection maps from 704 back to 1024 dimensions.  The output projection
is initialized to zero so that every residual expert initially preserves the
base representation.  We use a fixed residual scale of one.
The Null route is parameter-free and returns $\Delta h_0=0$.  The eight
experts account for approximately 17.30M parameters.  The linear router
contributes only approximately 9k additional parameters, giving about 17.31M
trainable parameters overall, reported as 17.3M (0.52\% of the complete VLA
policy) in the main paper.

\subsection{Local Success Prototypes}

For each task, we construct a fixed pool from chunk-level representations of
reliable successful deployment rollouts.  Successful rollouts that contain a
visually apparent intermediate deviation are excluded using the reliability
indicator $\gamma_i$.  Failed-rollout prefixes are used by Preserve but are
not included in the prototype pool.

Before retrieval, each pooled representation is $\ell_2$-normalized.  For a
failure-associated representation $h$, we retrieve its
$K_{\mathrm{nn}}=20$ exact nearest neighbors from the corresponding
task-specific pool.  The local success prototype $s(h)$ is the
coordinate-wise median of the normalized neighbors.  We use the same
normalized representation space when evaluating the distance from $h$ or a
candidate $c_k=h+\Delta h_k$ to $s(h)$.


\begin{table*}[!t]
\refstepcounter{algorithm}
\label{supp:alg:failpatch_training}
\centering
\scriptsize
\begin{tabular}{rp{15.2cm}}
\toprule
\multicolumn{2}{l}{\textbf{Algorithm \thealgorithm: FailPatch offline
preparation, joint training, and inference}} \\
\midrule
\multicolumn{2}{l}{\textbf{Input:} frozen policy $F_\phi$ and action
projection $W_v$; demonstrations $\mathcal{D}_{\mathrm{demo}}$; deployment
rollouts $\mathcal{B}$; $K$ experts $\{f_k\}_{k=1}^{K}$; router $g_\theta$.}
\\
\midrule
1 & Given the terminal outcome $y_i$, annotate each rollout $i\in\mathcal{B}$
to obtain failure onset $\widehat{t}_i^{\mathrm{fail}}$ for failed rollouts
and success reliability $\gamma_i$ for successful rollouts. \\
2 & Cache $h_{i,t}=F_\phi(o_{i,t},\ell_i)$ for all deployment chunks.  Form
$\mathcal{H}^{+}$ from reliable-success chunks and failed-rollout pre-onset
chunks; form $\mathcal{H}^{-}$ from onset/post-onset chunks. \\
3 & For each task $\tau$, construct $\mathcal{P}_\tau$ from normalized chunks
of reliable successful rollouts only. \\
4 & \textbf{For} $n=1,\ldots,30{,}000$: \\
5 & \quad Sample task-balanced chunks for the demonstration, Preserve, and Redirect branches according to the implementation, without trajectory-duration reweighting. \\
6 & \quad \textbf{Demonstration branch:} run $F_\phi$ online; set $c_0=h$,
$c_k=h+f_k(h)$, and $\ell_k=\lVert W_vc_k-u\rVert_2^2$. \\
7 & \quad Compute detached $w_k\propto\exp(-\ell_k/T_d)$ and
$\mathcal{L}_{\mathrm{expert}}
=\mathbb{E}_{\mathcal{D}_{\mathrm{demo}}}
[\sum_{k=1}^{K}w_k\ell_k]$. \\
8 & \quad If $n\leq5{,}000$, disable Null and set
$q^{\mathrm{succ}}_0=0,\ q^{\mathrm{succ}}_k=w_k$; otherwise set
$q^{\mathrm{succ}}_0=\mathbb{I}[\min_{k\geq1}\ell_k\geq\ell_0]$ and
$q^{\mathrm{succ}}_k=(1-q^{\mathrm{succ}}_0)w_k$. \\
9 & \quad Compute
$\mathcal{L}_{\mathrm{router-demo}}
=\mathbb{E}_{\mathcal{D}_{\mathrm{demo}}}
[\mathrm{CE}_{\mathrm{bal}}
(\operatorname{sg}(q^{\mathrm{succ}}),p_\theta(h))]$. \\
10 & \quad \textbf{Preserve branch:}
$\mathcal{L}_{P}=\mathbb{E}_{\mathcal{H}^{+}}
[-\log p_{\theta,0}(h)+
\sum_{k=1}^{K}p_{\theta,k}(h)\lVert f_k(h)\rVert_2^2]$. \\
11 & \quad \textbf{Redirect branch:} for each $h\in\mathcal{H}^{-}$, retrieve
$K_{\mathrm{nn}}=20$ task-matched neighbors from $\mathcal{P}_\tau$ and let
$s(h)$ be their coordinate-wise median. \\
12 & \quad Compute
$a_k=d(h,s(h))-d(c_k,s(h))-\lambda_u\lVert f_k(h)\rVert_2^2$ and form the
detached target $q(h)$ using Null threshold $m$ and temperature $T$. \\
13 & \quad Compute
$\mathcal{L}_{R}=\mathbb{E}_{\mathcal{H}^{-}}
[-\sum_{j=0}^{K}q_j\log p_{\theta,j}
+\lambda_r\sum_{k=1}^{K}q_k
[m_r-(d(h,s)-d(c_k,s))]_{+}]$. \\
14 & \quad \textbf{Trust branch:}
$\mathcal{L}_{T}=\mathbb{E}_{\mathcal{H}^{-}}
[\sum_{k=1}^{K}q_k\lVert f_k(h)\rVert_2^2]$. \\
15 & \quad Minimize
$\mathcal{L}_{\mathrm{expert}}
+\lambda_{rd}\mathcal{L}_{\mathrm{router-demo}}
+\lambda_P\mathcal{L}_{P}
+\lambda_R\mathcal{L}_{R}
+\lambda_T\mathcal{L}_{T}$; clip the global gradient norm to 1.0 and update
only $\theta$ and $\{f_k\}_{k=1}^{K}$. \\
16 & \textbf{Inference:} set
$k^\star=\arg\max_{k\in\{0,\ldots,K\}}p_{\theta,k}(h)$; use $h$ if
$k^\star=0$, otherwise use $h+f_{k^\star}(h)$ before $W_v$. \\
\bottomrule
\end{tabular}
\end{table*}




\subsection{Optimization Details}

Reliable successful-rollout chunks and pre-onset chunks from failed rollouts provide representation-level supervision only and do not supply action targets. The failure-onset chunk and all subsequent chunks are assigned to $\mathcal{H}^{-}$.

The first 5k of the 30k optimization steps use a Null-disabled warm-up.  During
this period, the demonstration-derived router target distributes
responsibility only among the eight residual experts, preventing the
zero-initialized candidates from collapsing immediately to the Null route.
Preserve, Redirect, and Trust supervision is otherwise active from the start
of training.  After warm-up, the Null route is enabled whenever no residual
candidate improves on the frozen base candidate.

The demonstration path performs a live forward pass through the frozen
backbone and retains the default $\pi_{0.5}$ image augmentation.  Deployment
representations used by the Preserve and Redirect branches are cached after
annotation.  All soft responsibilities are detached.  We clip the global
gradient norm to 1.0 and clamp routing probabilities below by $10^{-6}$ when
computing log probabilities. Trust is applied only in the hidden representation space and penalizes
$\lVert\Delta h_k\rVert_2^2$.

Table~\ref{supp:tab:loss_hparams} summarizes the verified default settings.

\begin{table}[!t]
\centering
\small
\begin{tabular}{lc}
\toprule
Configuration & Value \\
\midrule
Number of residual experts $K$ & 8 \\
Nearest neighbors $K_{\mathrm{nn}}$ & 20 \\
Demo routing weight $\lambda_{rd}$ & 1.0 \\
Preserve weight $\lambda_P$ & 0.2 \\
Redirect weight $\lambda_R$ & 0.2 \\
Trust weight $\lambda_T$ & 0.05 \\
Utility residual penalty $\lambda_u$ & 0.01 \\
Failure routing temperature $T$ & 0.2 \\
Null utility threshold $m$ & 0.1 \\
Redirect margin weight $\lambda_r$ & 0.01 \\
Redirect margin $m_r$ & 0.1 \\
$T_d$ schedule & $1.0\rightarrow0.2$ \\
Null-disabled warm-up & 5k steps \\
Total optimization budget & 30k steps \\
\bottomrule
\end{tabular}
\caption{Default FailPatch training configuration.}
\label{supp:tab:loss_hparams}
\end{table}

\subsection{Training and Inference Procedure}

Algorithm~\ref{supp:alg:failpatch_training} details the joint training step.  The
three data branches are optimized together rather than in separate stages.

The offline annotator, trajectory labels, cached rollout features, and
prototype pools are discarded after training.


\begin{figure*}[htbp]
\centering
\begin{minipage}{0.95\textwidth}

\noindent\textbf{System prompt.}
\begin{lstlisting}[style=vlmprompt]
You are an expert robotic manipulation analyst. Your job is to analyze temporally ordered RGB observations from a dual-arm robot episode.

For a failed rollout, identify the earliest action chunk at which a visible and actionable mistake or wrong commitment begins, marking the failure onset before the episode becomes irreversible.

For a successful rollout, determine whether it is a reliable success. A rollout is reliable only if it contains no visually apparent substantive intermediate deviation, such as a failed grasp, collision, incorrect placement, wrong object selection, prolonged stall, or retry.

Respond only with valid JSON matching the requested schema. If uncertain, return the best estimate and set confidence accordingly.
\end{lstlisting}

\vspace{3pt}

\noindent\textbf{User prompt template.}
\begin{lstlisting}[style=vlmprompt]
Task instruction: {task_instruction}

Rollout outcome: {rollout_outcome}

You are given a temporally ordered contact sheet of RGB frames from a single deployment rollout. Each tile is labeled with chunk index t=0,1,2,... in execution order. One frame per action chunk is shown (head camera unless noted).

Definitions:

failure_onset_chunk: for a failed rollout, the earliest chunk where a visible mistake or wrong commitment appears that will likely cause final task failure if not corrected.
reliable_success: for a successful rollout, whether the trajectory contains no visually apparent substantive intermediate deviation that would make it unreliable as a success prototype (true = reliable, false = unreliable).

Rules:

If the rollout outcome is failure, mark the earliest visible wrong action or commitment, not the moment when failure is already irreversible.
If the rollout outcome is success, set failure_onset_chunk and failure_type to null, and determine reliable_success based on the complete trajectory.
A successful rollout with a failed grasp, visible collision, incorrect placement, wrong object selection, prolonged stall, or retry should be marked as reliable_success=false, even if the task is eventually completed.
Short-lived motion jitter that does not disrupt task progress is not sufficient to mark a successful rollout as unreliable.
If multiple errors occur in a failed rollout, return the first visible error.
If RGB observations are insufficient, return null for the corresponding uncertain field and explain the reason in notes.
Chunk indices are 0-based and align with policy action chunks (50 environment steps per chunk in simulation).

Output JSON schema:
{
"failure_onset_chunk": <int or null>,
"confidence": <float 0-1>,
"failure_type": <string or null>,
"notes": <string>,
"reliable_success": <bool or null>
}

For a failed rollout, set reliable_success to null.
For a successful rollout, set failure_onset_chunk and failure_type to null.

Return JSON only.
\end{lstlisting}

\end{minipage}
\caption{System and user prompts used for VLM-based rollout annotation.}
\label{supp:fig:vlm-prompt}
\end{figure*}

\section{Offline VLM Annotation}
\label{supp:sec:annotation}

\subsection{Input and Output Format}

We use GPT-5.5 as an offline
annotator. The annotator is never used during policy execution.  This use
of a multimodal model for offline failure reasoning is related to prior
VLM-based failure analysis~\citep{duan2025aha}, but our output is restricted
to a chunk-level onset and a success-reliability label.

For each rollout, we construct a temporally ordered contact sheet from the
head-camera RGB stream.  We include one frame per action chunk when the
episode contains at most 24 chunks.
Each image is resized so that its shorter side is 224 pixels.  The contact sheet labels every frame with its original
chunk index.  The task instruction and the corresponding terminal outcome label accompany the images.


For every rollout, the annotator returns failure\_onset\_chunk, confidence, failure\_type, notes, and reliable\_success. For failed rollouts, reliable\_success is set to null; for successful rollouts, failure\_onset\_chunk and failure\_type are set to null.

\subsection{Verbatim Annotation Prompt}

\paragraph{Prompt roles.}
The \emph{system prompt} is fixed across all rollouts and defines the
annotator's role, the failure-onset localization and reliable-success assessment objectives, and the requirement to return
valid JSON.  The user prompt template is instantiated for each rollout:
\texttt{task\_instruction} is replaced by the corresponding task
instruction, \texttt{rollout\_outcome} is replaced by the terminal
outcome label (\texttt{success} or \texttt{failure}), and the temporally
ordered contact sheet is supplied as the image input.
Figure~\ref{supp:fig:vlm-prompt} reproduces both prompts verbatim.



\subsection{Reliable-Success Filtering}

A successful rollout is assigned $\gamma_i=0$ if it contains a substantive
intermediate deviation, such as selecting an incorrect object and later
switching back, a visible collision, a failed placement followed by a retry,
or a long execution stall followed by recovery.  Short-lived motion jitter is
not sufficient for exclusion.  Rollouts with $\gamma_i=0$ are excluded from
the reliable prototype and Preserve pools.  Table~\ref{supp:tab:gamma_filter}
reports the retained and excluded successful rollouts for each task.

\begin{table}[h]
\centering
\resizebox{\columnwidth}{!}{
\begin{tabular}{llrrr}
\toprule
Setting & Task & Successful rollouts & $\gamma=1$ & $\gamma=0$ \\
\midrule
Simulation & Blocks Ranking & 50 & 44 & 6 \\
Simulation & Stack Bowls & 50 & 46 & 4 \\
Simulation & Put Object Cabinet & 50 & 43 & 7 \\
Simulation & Place Bread Basket & 50 & 45 & 5 \\
\midrule
Real world & Blocks Ranking & 50 & 44 & 6 \\
Real world & Stack Bowls & 50 & 45 & 5 \\
Real world & Fold Dishcloth & 50 & 43 & 7 \\
\bottomrule
\end{tabular}
}
\caption{Reliable-success filtering statistics.}
\label{supp:tab:gamma_filter}
\end{table}

\subsection{Human Verification}

The VLM onset lies within one action chunk of the human label for 160 of 200
rollouts (80\%) and within two chunks for 190 rollouts (95\%).  The mean
absolute error is 0.74 chunks, with a mean signed error of $+0.14$ chunks,
indicating a small tendency to predict the onset later than the human label.
Table~\ref{supp:tab:annotation_agreement} provides the task-level agreement
statistics.

The Human-Annotated Onset ablation replaces only the failed-rollout onset
labels; it retains GPT-5.5 reliable-success filtering and all other training
and evaluation settings.  Its 61.00\% average success rate is close to the
60.50\% obtained with GPT-5.5 onsets, indicating that the modest localization
error has little effect under the present training protocol.

\begin{table}[htbp]
\centering
\scriptsize
\resizebox{\columnwidth}{!}{%
\begin{tabular}{lrrrrrr}
\toprule
Task & $N$ & Exact & $|\Delta|\leq1$ & $|\Delta|\leq2$ & MAE & Bias \\
\midrule
Blocks Ranking & 50 & 28 & 40 & 47 & 0.72 & $+0.14$ \\
Stack Bowls & 50 & 30 & 42 & 48 & 0.68 & $+0.10$ \\
Put Object Cabinet & 50 & 26 & 39 & 48 & 0.80 & $+0.20$ \\
Place Bread Basket & 50 & 28 & 39 & 47 & 0.76 & $+0.12$ \\
\midrule
Overall & 200 & 112 & 160 & 190 & 0.74 & $+0.14$ \\
\bottomrule
\end{tabular}
}
\caption{Agreement between GPT-5.5 and human failure-onset annotations. Errors
are measured in action chunks.}
\label{supp:tab:annotation_agreement}
\end{table}

\section{Experimental Details}
\label{supp:sec:experiments}

\begin{figure*}[!t]
\centering
\includegraphics[width=0.98\textwidth]
{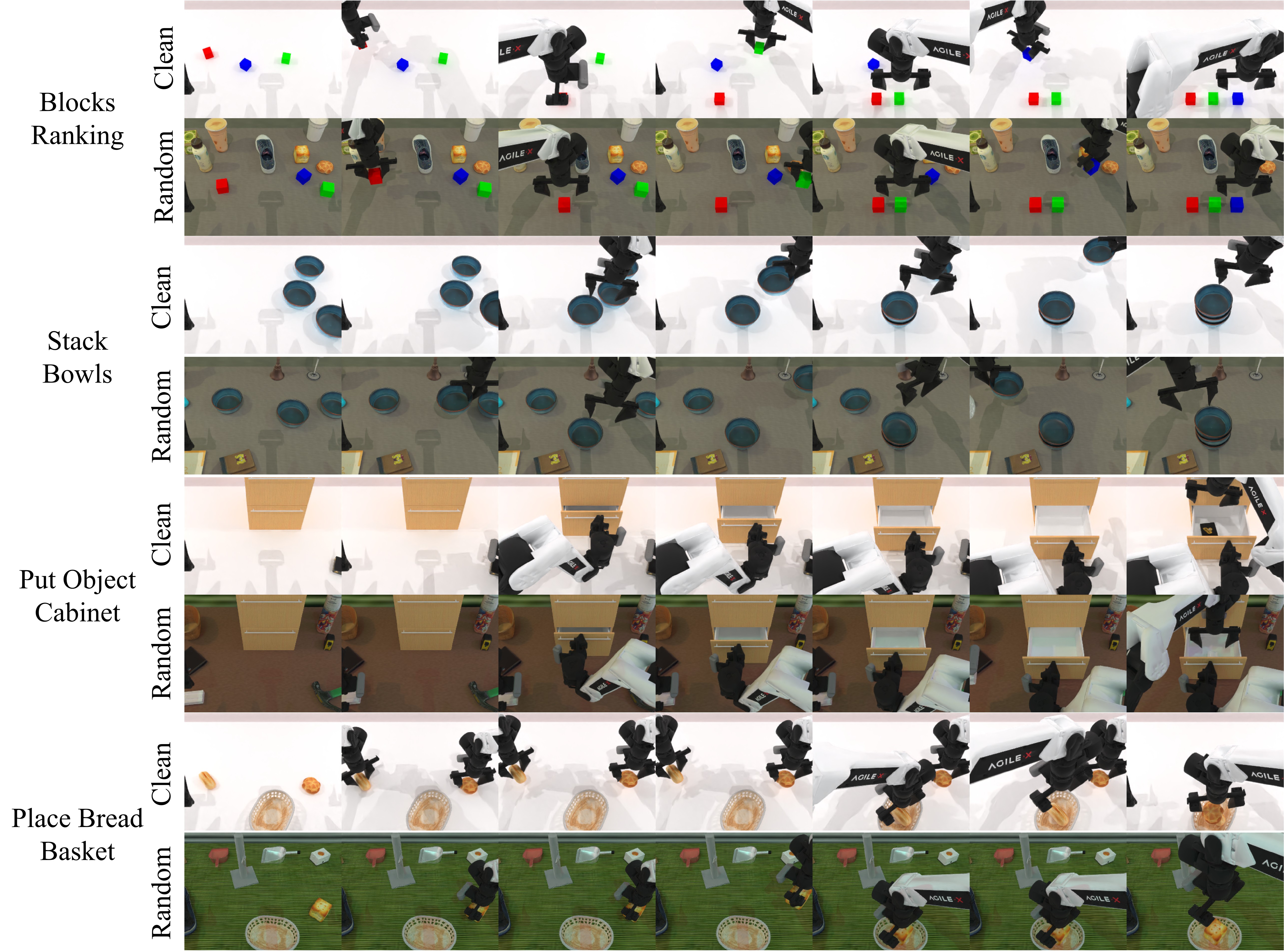}
\vspace{-6pt}
\caption{Representative execution sequences under clean and random evaluation
for Blocks Ranking, Stack Bowls, Put Object Cabinet, and Place Bread Basket.
Object pose and identity vary in both settings, while the random setting also
changes background textures, clutter, table height, and lighting.}
\label{supp:fig:sim_show}
\end{figure*}

\subsection{Simulation Tasks}

We use RoboTwin 2.0 \citep{chen2025robotwin} with a SAPIEN
 \citep{Xiang_2020_SAPIEN} backend.  Each policy prediction contains 50
environment actions, so the 1200-step tasks admit at most 24 chunks and the
700-step tasks admit at most 14 chunks.  Table~\ref{supp:tab:sim_tasks} gives the
exact code-level success predicates; success is checked after every
environment step and terminates the episode immediately.

\begin{table*}[!t]
\centering
\small
\begin{tabular}{lccp{10.0cm}}
\toprule
Task & Max steps & Max chunks & Success predicate \\
\midrule
Blocks Ranking
& 1200 & 24
&For the two adjacent block pairs (red, green) and (green, blue), $|\Delta x|<0.13$\,m and
$|\Delta y|<0.03$\,m; the centers satisfy
$x_{\mathrm{red}}<x_{\mathrm{green}}<x_{\mathrm{blue}}$; and both grippers
are open. \\
Stack Bowls
& 1200 & 24
& After sorting the bowls by height, every pair has planar center distance
below 0.04\,m; their heights are within 0.02\,m of
$[0.74,0.77,0.81]+\texttt{table\_z\_bias}$; and both grippers are open. \\
Put Object Cabinet
& 700 & 14
& The planar distance from the object to the drawer functional point is below
0.05\,m; its relative height satisfies
$0.007<\Delta z<0.12$\,m; and the executing gripper is open. \\
Place Bread Basket
& 700 & 14
& Every bread object's planar distance to the basket center is below 0.05\,m,
its height exceeds $0.73+\texttt{table\_z\_bias}$, and both grippers are
open. \\
\bottomrule
\end{tabular}
\caption{Simulation horizons and exact success predicates.}
\label{supp:tab:sim_tasks}
\end{table*}

\subsection{Per-Seed Object Randomization}

Object pose, scale, and identity are sampled from the task distributions in
Table~\ref{supp:tab:object_randomization} for \emph{both} clean and random evaluation.  


\begin{table*}[!t]
\centering
\small
\begin{tabular}{lp{13.6cm}}
\toprule
Task & Randomized task instance \\
\midrule
Blocks Ranking
& Each block is initialized with $x\in[-0.28,0.28]$\,m,
$y\in[-0.08,0.05]$\,m, $z=0.765$\,m, and rotation about the vertical axis up
to 0.75\,rad.  The block side length is sampled uniformly from
$[0.03,0.05]$\,m.  Target $x$ ranges are
$[-0.09,-0.08]$, $[-0.01,0.01]$, and $[0.08,0.09]$\,m for red, green, and
blue, respectively, with common target $y\in[-0.20,-0.10]$\,m. \\
Stack Bowls
& Bowl centers are initialized with $x\in[-0.30,0.30]$\,m and
$y\in[-0.15,0.15]$\,m.  Bowl identities are
assigned according to their initial $y$ ordering, and the target stack center
is $(0,-0.10)$\,m. \\
Put Object Cabinet
& The cabinet has $x\in[-0.05,0.05]$\,m and $y=0.155$\,m.  The manipulated
object and model instance are sampled from ten object categories.  The object
has $x\in[-0.25,0.25]$\,m, $y\in[-0.20,-0.10]$\,m, and yaw up to
$\pi/3$; when the initially sampled $|x|<0.20$\,m, the implementation widens
the admissible $x$ range to $[-0.32,0.32]$\,m. \\
Place Bread Basket
& The basket center is fixed at $(0,-0.20)$\,m, while its yaw is sampled from
$[0,\pi]$ and its model ID from $\{0,1,2,3,4\}$.  Each instance contains one
or two bread objects with $x\in[-0.27,0.27]$\,m,
$y\in[-0.20,0.05]$\,m, yaw up to $\pi/4$, and model ID sampled from
$\{0,1,3,5,6\}$. \\
\bottomrule
\end{tabular}
\caption{Task-instance randomization applied independently for every seed in
both clean and random evaluation.}
\label{supp:tab:object_randomization}
\end{table*}

\subsection{Clean-to-Random Domain Shift}

All demonstrations and fixed deployment buffers are collected under
\texttt{demo\_clean.yml}.  Clean-to-random evaluation switches only to
\texttt{demo\_randomized.yml} and reuses the same adapted checkpoint.  As
Table~\ref{supp:tab:domain_randomization} shows, the shift adds visual and scene
randomization on top of the per-seed task-instance randomization above; the
camera configuration is unchanged.

\begin{table*}[!t]
\centering
\small
\begin{tabular}{p{3.6cm}p{5.0cm}p{7.0cm}}
\toprule
Attribute & Clean & Random \\
\midrule
Background
& Default wall and table textures; clean-background probability 1.0.
& Unseen evaluation texture library; clean-background probability 0.02. \\
Table clutter
& Disabled.
& Enabled, with up to approximately ten additional distractor objects. \\
Table height
& No perturbation.
& Downward height perturbation of up to 0.03\,m. \\
Lighting
& Fixed.
& Randomized; extreme lighting is sampled with probability 0.02. \\
Head-camera distance
& No distance perturbation.
& No distance perturbation. \\
Camera observations
& One D435 head view and two wrist views.
& Identical camera configuration. \\
\bottomrule
\end{tabular}
\caption{Difference between the clean and random simulation configurations.
Object pose/type randomization is active in both settings.}
\label{supp:tab:domain_randomization}
\end{table*}

\begin{table*}[!t]
\centering
\small
\begin{tabular}{lrrrrrr}
\toprule
Method & Trainable params & Fraction & Time & Offline rollouts/task &
Fresh rollouts \\
\midrule
LoRA SFT & 50.0M & 1.49\% & 8.3 h & 0 & 0 \\
Full SFT & 3353.4M & 100\% & 11.2 h & 0 & 0 \\
Demo-Only REB & 17.3M & 0.52\% & 3.1 h & 0 & 0 \\
Outcome-Prompt SFT (Episode) & 17.3M & 0.52\% & 5.5 h & 100 & 0 \\
Outcome-Prompt SFT (Chunk) & 17.3M & 0.52\% & 4.7 h & 100 & 0 \\
PPO & 694M & 20.7\% & 20.8 h & 0 & 12,800 \\
FailPatch & 17.3M & 0.52\% & 3.1 h & 100 & 0 \\
\bottomrule
\end{tabular}
\caption{Trainable parameters, wall-clock adaptation time, and rollout usage.
The PPO rollout count is the total over 100 updates.}
\label{supp:tab:efficiency}
\end{table*}

Figure~\ref{supp:fig:sim_show} further illustrates representative execution
sequences for all four tasks under both clean and random evaluation.

\subsection{Evaluation Protocol}

For each task, we use environment seeds $\{100000,\ldots,100099\}$ for every method. Each reported simulation result is the success rate over these same 100 initial-state seeds.

\subsection{Supervised Baselines}

All supervised adaptation methods---LoRA SFT, Full SFT, Demo-Only REB, both
Outcome-Prompt variants, and FailPatch---reuse the same demonstrations, action
normalization statistics, augmentation pipeline, and optimization
configuration.  Each is trained for 30k steps with a peak learning rate of
$2.5\times10^{-5}$, weight decay of $10^{-10}$, and a global batch size of
32.  We clip the gradient norm to 1.0, use
training seed 42, and evaluate the final 30k-step checkpoint.
LoRA SFT~\citep{hu2022lora} inserts rank-64 adapters with scale 128 and
dropout 0.05 into the
language and action-expert attention and feed-forward projections, resulting
in approximately 50.0M trainable parameters.  Full SFT updates the complete
VLA model---the vision encoder, language model, action expert, projection
layers, and time MLP---for approximately 3.35B trainable parameters.
Demo-Only REB uses the same router and residual-expert architecture as
FailPatch but receives no deployment-trajectory supervision.

The two outcome-conditioned SFT baselines use identical adaptation data and
parameter budgets.  They append the literal condition
\texttt{success: True/False} to the task instruction.  The episode-level
variant assigns the terminal outcome to every action chunk.  The chunk-level
variant uses temporally localized reliability labels: failed-rollout chunks
before the onset receive \texttt{success: True}, whereas the onset and
subsequent chunks receive \texttt{success: False}.  Both variants are
evaluated under the desired condition \texttt{success: True}.


\subsection{PPO Baseline}

PPO~\citep{schulman2017proximal} is not part of the supervised configuration above
and is optimized separately. It starts from the same 50-Task Base and uses a sparse
terminal success reward. The visual and language backbones remain frozen
(\texttt{train\_expert\_only=True}), while PPO updates the 694M trainable
parameters (action expert).
Training runs for 100 policy updates; each update collects 128 fresh on-policy
rollouts.
We use discount factor $\gamma=0.99$, GAE parameter $\lambda=0.95$, PPO clip
ratio 0.2, dual-clip ratio 3.0, and value clipping 0.2. Advantages are
normalized within each update. We perform five optimization epochs per policy
update with a global batch size of 1{,}024 (micro-batch size 32 per GPU on
four GPUs). The policy and value learning rates are $5\times10^{-6}$ and
$10^{-4}$, respectively. We do not add an entropy bonus or KL penalty
(entropy\_bonus=0, kl\_beta=0). During training rollouts,
actions are sampled with flow-SDE exploration noise\_level=0.3;
evaluation uses deterministic flow-ODE inference. PPO uses the same evaluation
seeds as the supervised method.

\subsection{Compute and Runtime}

All adaptation methods are trained on four NVIDIA H20 GPUs.  Times in Table
\ref{supp:tab:efficiency} are wall-clock training times.  FailPatch training excludes
the one-time offline VLM annotation and rollout-feature caching cost.  It
requires no environment interaction after its fixed deployment buffer has
been collected.

\begin{table*}[!t]
\centering
\small
\begin{tabular}{lrrrrr}
\toprule
Prototype variant & Blocks & Stack & Cabinet & Bread & Average \\
\midrule
Default: $K_{\mathrm{nn}}=20$, median & 54 & 77 & 49 & 62 & \textbf{60.50} \\
$K_{\mathrm{nn}}=1$ & 48 & 71 & 44 & 58 & 55.25 \\
$K_{\mathrm{nn}}=10$ & 53 & 76 & 48 & 61 & 59.50 \\
$K_{\mathrm{nn}}=50$ & 52 & 75 & 47 & 61 & 58.75 \\
Mean aggregation & 52 & 75 & 47 & 60 & 58.50 \\
Unnormalized features & 51 & 74 & 46 & 60 & 57.75 \\
Task-level global prototype & 50 & 73 & 45 & 59 & 56.75 \\
\bottomrule
\end{tabular}
\caption{Prototype-construction ablations under clean simulation evaluation.
The default uses a task-specific reliable-success pool, normalized features,
Euclidean retrieval, and coordinate-wise median aggregation.}
\label{supp:tab:prototype_ablation}
\end{table*}

\begin{table*}[!t]
\centering
\small
\begin{tabular}{llll}
\toprule
Hyperparameter & Evaluated values & Default & Average success (\%) \\
\midrule
Experts $K$ & 4, 8, 16 & 8 & 54.25, \textbf{60.50}, 58.00 \\
Preserve weight $\lambda_P$ & 0.1, 0.2, 0.4 & 0.2
& 57.50, \textbf{60.50}, 58.25 \\
Redirect weight $\lambda_R$ & 0.1, 0.2, 0.4 & 0.2
& 56.00, \textbf{60.50}, 59.25 \\
Trust weight $\lambda_T$ & 0.02, 0.05, 0.10 & 0.05
& 58.75, \textbf{60.50}, 57.00 \\
Demo temperature $T_d$ & annealed, 0.2, 1.0 & annealed
& \textbf{60.50}, 59.00, 55.50 \\
Failure temperature $T$ & 0.1, 0.2, 0.5 & 0.2
& 59.25, \textbf{60.50}, 57.75 \\
Null threshold $m$ & 0.05, 0.1, 0.2 & 0.1
& 58.50, \textbf{60.50}, 57.25 \\
Redirect margin $m_r$ & 0.05, 0.1, 0.2 & 0.1
& 59.00, \textbf{60.50}, 58.00 \\
\bottomrule
\end{tabular}
\caption{One-dimensional hyperparameter sensitivity under clean simulation
evaluation.}
\label{supp:tab:sensitivity}
\end{table*}

\begin{table*}[!t]
\centering
\small
\begin{tabular}{lrrrrr}
\toprule
Variant & Blocks & Stack & Cabinet & Bread & Average \\
\midrule
Demo-Only REB & 42 & 64 & 36 & 55 & 49.25 \\
$+$ Preserve & 50 & 63 & 38 & 59 & 52.50 \\
$+$ Redirect & 52 & 68 & 42 & 64 & 56.50 \\
$+$ Trust & 44 & 62 & 37 & 58 & 50.25 \\
$+$ Preserve $+$ Redirect & 54 & 74 & 48 & 60 & 59.00 \\
\midrule
Parameter-Matched Single Expert & 50 & 71 & 44 & 58 & 55.75 \\
REB without Null Expert & 51 & 68 & 47 & 60 & 56.50 \\
Human-Annotated Onset & 54 & 78 & 50 & 62 & 61.00 \\
Full FailPatch & 54 & 77 & 49 & 62 & 60.50 \\
\bottomrule
\end{tabular}
\caption{Per-task objective, architecture, and annotation ablations under
clean simulation evaluation.}
\label{supp:tab:full_ablation}
\end{table*}

\section{Additional Ablation Studies}
\label{supp:sec:ablations}

\subsection{Prototype Construction}

The default method constructs a query-dependent local success prototype for
each failure-associated representation. Specifically, it retrieves the
$K_{\mathrm{nn}}=20$ nearest representations from the task-specific pool of
$\ell_2$-normalized reliable successful-rollout representations and takes
their coordinate-wise median.

We compare several alternative constructions. The
$K_{\mathrm{nn}}\in\{1,10,50\}$ variants change only the number of retrieved
neighbors. Mean aggregation replaces the coordinate-wise median of the same
20 neighbors with their coordinate-wise mean. Unnormalized features performs retrieval and prototype
construction in the original feature space without $\ell_2$ normalization.
Finally, the task-level global prototype replaces the query-dependent local
prototype with a single prototype computed from the entire task-specific
reliable-success pool. Each variant changes only one component of the default
configuration. Table~\ref{supp:tab:prototype_ablation} reports the complete
comparison.

\subsection{Hyperparameter Sensitivity}

We vary one hyperparameter at a time while retaining the remaining Full
FailPatch defaults.  Table~\ref{supp:tab:sensitivity} shows that the default
configuration performs best among the evaluated values.

\subsection{Per-Task Objective and Architecture Ablations}

Table~\ref{supp:tab:full_ablation} expands the average results in the main paper.
Redirect provides the largest standalone improvement over Demo-Only REB, and
Preserve further complements it.  A parameter-matched single expert and an
expert bank without a Null route both underperform the complete model,
supporting diverse intervention directions and learned abstention.

\subsection{Failure-Data Efficiency}

\begin{figure}[t]
\centering
\includegraphics[width=\columnwidth]{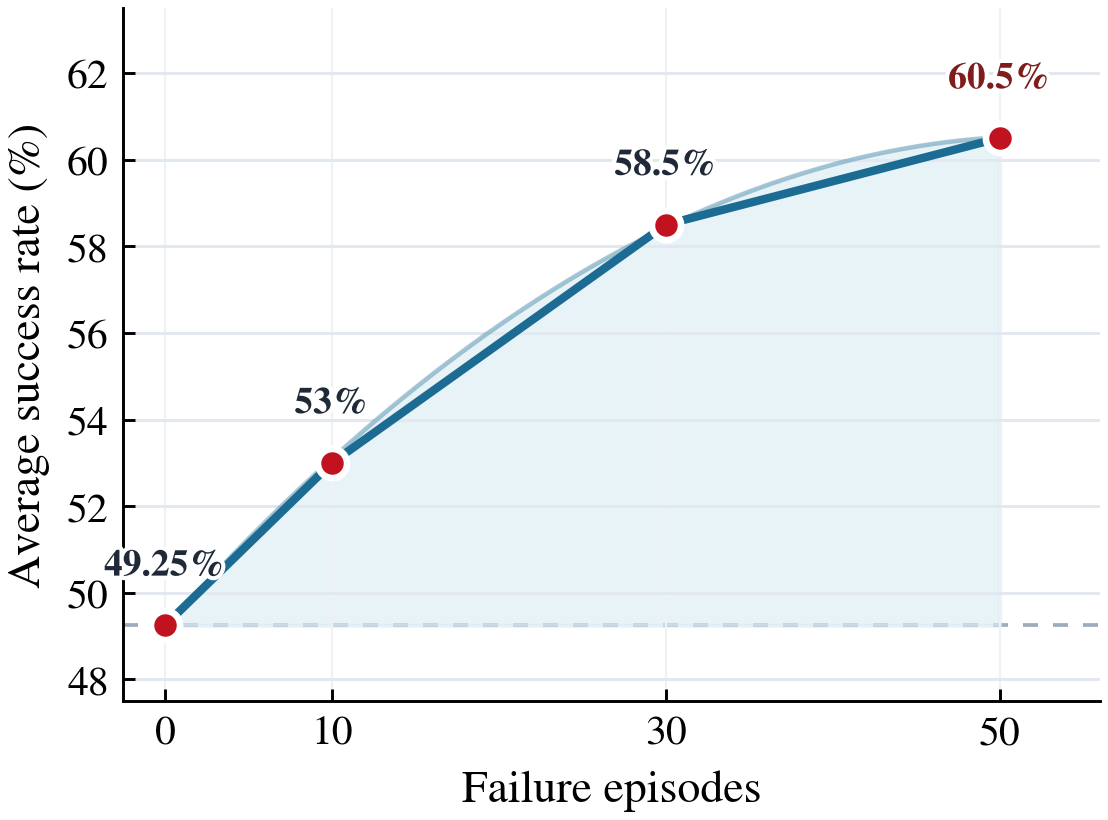}
\caption{Effect of failure-data scale on the average success rate across the four simulation tasks. The number of failed deployment episodes is varied independently for each task, and the 50-episode result corresponds to the full FailPatch setting.}
\label{supp:fig:failure_data_efficiency}
\end{figure}

We examine how the amount of failure data affects adaptation by varying the number of failed deployment episodes per task as $N_f\in\{0,10,30,50\}$. Across all settings, we keep the base policy, demonstration set, successful-rollout portion of the fixed deployment buffer, training budget, and evaluation protocol unchanged; only the number of failed rollouts is varied. For $N_f<50$, the corresponding subset is taken from the same fixed buffer of 50 failed rollouts per task without collecting additional trajectories. The zero-failure setting therefore retains supervision from reliable successful rollouts but removes all supervision derived from failed-rollout prefixes and onset/post-onset chunks. As shown in Fig.~\ref{supp:fig:failure_data_efficiency}, the average success rate across the four simulation tasks increases from 49.25\% with no failed rollouts to 53.0\%, 58.5\%, and 60.5\% with 10, 30, and 50 failed rollouts per task, respectively. Notably, using only 30 failed rollouts yields a 9.25-percentage-point improvement over the zero-failure setting and comes within 2.0 percentage points of the full-data result. These results demonstrate that FailPatch can learn useful residual corrections from a relatively modest failure buffer, with additional failure experience providing diminishing gains beyond 30 episodes per task.

\begin{table*}[htbp]
\centering
\small
\begin{tabular}{lp{7.0cm}p{6.1cm}}
\toprule
Task & Language instruction & Success criterion \\
\midrule
Blocks Ranking
& ``Position red block, green block, and blue block in a left-to-right
sequence, forming a row.''
& The three blocks form one row in red--green--blue order from left to right.
\\
Stack Bowls
& ``Stack the three bowls on the top of each other.''
& The task ends successfully once all three bowls have been stacked; no
additional fixed-duration stability test is imposed.
\\
Fold Dishcloth
& ``Place the dishcloth in the center of the table, then fold it in half
twice.''
& The dishcloth is first placed at the table center and then successfully
folded in half twice.
\\
\bottomrule
\end{tabular}
\caption{Language instructions and success criteria for the real-world tasks.}
\label{supp:tab:real_tasks}
\end{table*}

\begin{table}[htbp]
\centering
\small
\begin{tabular}{lcc}
\toprule
Feature & Balanced accuracy & ROC-AUC \\
\midrule
Vision & 0.73 & 0.79 \\
Action hidden & \textbf{0.74} & \textbf{0.83} \\
Predicted action & 0.62 & 0.69 \\
Shuffled labels & 0.51 & 0.52 \\
\bottomrule
\end{tabular}
\caption{Aggregate linear-probe results.}
\label{supp:tab:linear_probe}
\end{table}

\section{Additional Mechanistic Analysis}
\label{supp:sec:analysis}

\subsection{Linear-Probe Protocol}

We form the nominal class from pre-onset chunks of failed rollouts and the
failure-associated class from the localized onset and post-onset chunks of
the same rollouts.  We split at the rollout level using an
80/10/10 train/validation/test ratio so that chunks from the same trajectory
never appear in different partitions.

For each feature space, we fit an $\ell_2$-regularized logistic regression
probe with $C=1.0$ after fitting a StandardScaler on the training split.
Predicted actions are represented by flattening the complete $50\times32$
action chunk. The confidence intervals shown in Fig.~5 of the main paper are
95\% bootstrap intervals based on 1,000 rollout-level resamples.  Shuffled-label controls are repeated ten times.
Table~\ref{supp:tab:linear_probe} reports the aggregate test performance.

\section{Real-World Experimental Details}
\label{supp:sec:real}

\subsection{Platform and Control}

\begin{figure}[h]
\centering
\includegraphics[width=0.85\columnwidth]
{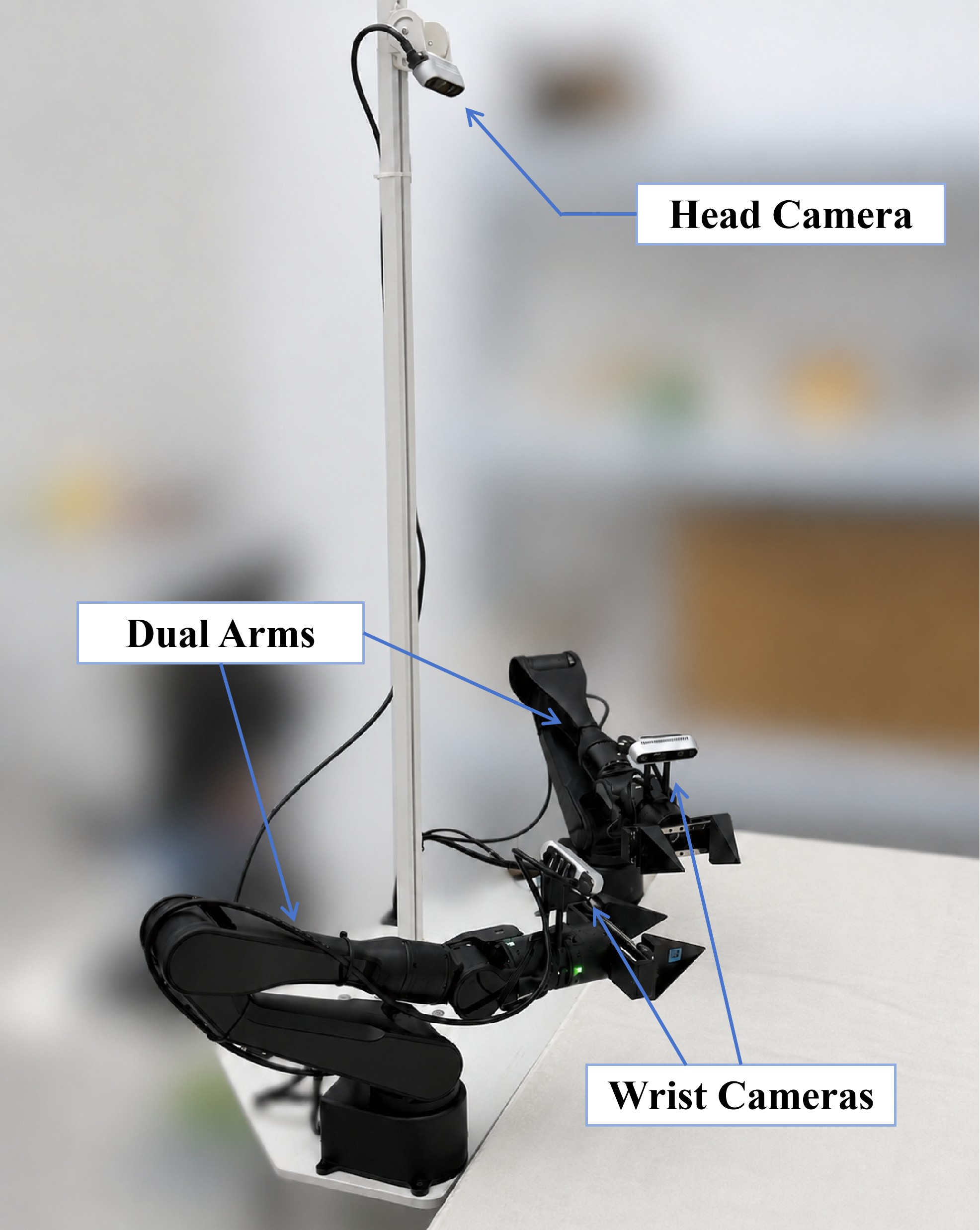}
\vspace{-6pt}
\caption{Real-robot platform.  The annotations identify the head-mounted
D435i, dual arms, and two wrist cameras.}
\label{supp:fig:real_platform}
\end{figure}

The physical platform comprises two six-degree-of-freedom robot arms with
parallel-jaw grippers.  Visual observations are provided by one head-mounted
Intel RealSense D435i and two wrist-mounted RGB cameras.  The policy predicts a 50-step chunk of dual-arm joint-space actions and gripper commands, executes the first 25 steps, and then replans from updated observations. Execution is subject to joint limits,
velocity clipping, collision monitoring, and a hardware emergency stop.
Inference runs on one RTX 4090 GPU.  Figure~\ref{supp:fig:real_platform} shows the physical
setup and camera placement.

\subsection{Tasks and Success Criteria}

Table~\ref{supp:tab:real_tasks} lists the exact language instructions and success
criteria used for the three physical tasks.

No human intervention is permitted after a trial begins.  Each trial is
independently reset.  For every task, a single operator collects 100 demonstrations.  These 300 demonstrations jointly train the Three-Task
Base and are reused for adaptation.  We additionally collect rollouts from
the frozen base policy until 50 successful and 50 failed trajectories are
available per task. The same GPT-5.5 head-camera annotation procedure and the same
FailPatch hyperparameters are used in simulation and on the physical system.

Evaluation uses 30 independent trials per task and method.  The methods are
evaluated in separate blocks rather than randomly interleaved.  Every trial begins from a fresh task reset.  We use
a fixed 30k-step checkpoint.


\section{Limitations}

The main limitation of FailPatch is its reliance on an offline VLM to
localize failure onset from RGB observations. Although the
human-annotated-onset ablation suggests that modest localization errors have
limited impact on performance, failure onset may remain ambiguous when the
causal error is occluded or depends on unobserved force and contact
information. Future work could incorporate proprioceptive or force signals,
or develop self-supervised and uncertainty-aware localization, to reduce this
dependence.

\end{document}